%% file: main.tex
\documentclass[10pt]{article}

\usepackage[letterpaper,top=1in,bottom=1in,left=1in,right=1in]{geometry}
\usepackage[T1]{fontenc}
\usepackage[utf8]{inputenc}
\usepackage{newtxtext,newtxmath}
\usepackage{microtype}
\usepackage{amsmath,mathtools}
\usepackage{graphicx}
\usepackage{booktabs}
\usepackage{array,tabularx}
\usepackage{multirow}
\usepackage{enumitem}
\usepackage[numbers,sort&compress]{natbib}
\usepackage{xcolor}
\usepackage{tikz}
\usetikzlibrary{arrows.meta,positioning,calc,fit,shapes.geometric}
\usepackage{pgfplots}
\pgfplotsset{compat=1.18}
\usepgfplotslibrary{groupplots}
\usepackage{caption}
\usepackage{subcaption}
\usepackage{algorithm}
\usepackage[noend]{algpseudocode}
\usepackage{listings}
\usepackage{url}
\usepackage{hyperref}
\usepackage[nameinlink,capitalize,noabbrev]{cleveref}
\usepackage{float}

\usepackage{xspace}

\definecolor{RSIBlue}{HTML}{2563A6}
\definecolor{RSIPurple}{HTML}{7251B5}
\definecolor{RSIGreen}{HTML}{2A8C78}
\definecolor{RSIOrange}{HTML}{D97A2B}
\definecolor{RSIRed}{HTML}{B33A3A}
\definecolor{RSIInk}{HTML}{202A36}
\definecolor{RSIGray}{HTML}{F3F5F7}

\hypersetup{
  colorlinks=true, linkcolor=RSIBlue, citecolor=RSIPurple, urlcolor=RSIGreen,
  pdfauthor={Authors to be inserted},
  pdftitle={Compile Once, Run Cheap: Frozen, Portable Skills for Embodied Agents, and What Recursive Self-Improvement Adds}
}

\setlist[itemize]{leftmargin=*,topsep=2pt,itemsep=1.5pt,parsep=0pt}
\setlist[enumerate]{leftmargin=*,topsep=2pt,itemsep=1.5pt,parsep=0pt}
\pgfplotsset{
  discard if not/.style 2 args={
    x filter/.append code={
      \edef\tempa{\thisrow{#1}}
      \edef\tempb{#2}
      \ifx\tempa\tempb\else\fi
    }
  }
}

\newcommand{\method}{\textsc{Skill-RSI}\xspace}
\newcommand{\fs}{\textsc{FrozenSkill}\xspace}
\newcommand{\protoA}{Protocol~A\xspace}
\newcommand{\protoB}{Protocol~B\xspace}

\input{data/results_numbers}
\input{data/tab_compile_rows}
\input{data/tab_extra_eval_rows}
\input{data/tab_heldout_rows}
\input{data/tab_writeonce_rows}
\input{data/tab_reuse_rows}
\input{data/tab_f08_rows}
\input{data/tab_f08_heldout_rows}
\input{data/tab_f08_compare_rows}
\input{data/tab_f09_rows}

\lstdefinestyle{skill}{
  basicstyle=\ttfamily\scriptsize, backgroundcolor=\color{RSIGray}, frame=single,
  rulecolor=\color{RSIInk!22}, framerule=0.4pt, breaklines=true, columns=fullflexible,
  keepspaces=true, xleftmargin=2pt, xrightmargin=2pt, aboveskip=4pt, belowskip=4pt, showstringspaces=false
}

\begin{document}

\begin{center}
  {\LARGE\bfseries Compile Once, Run Cheap:\\ Frozen, Portable Skills for Embodied Agents,\\ and What Recursive Self-Improvement Adds\par}
  \vspace{0.55em}
  {\large Jingwen Bai$^{1}$, Guochao Bai$^{2}$, Canbo Zhang$^{1}$, Shuangqi Li$^{3}$, Minchao Hu$^{3}$, Weichun Huang$^{2}$, Shuang Liu$^{1,*}$\par}
  \vspace{0.4em}
  {\small $^{1}$Bei{}jing Ou Rui Ling Shi Technology Co., Ltd.\\
  $^{2}$ChangingTek Robotics Technology (Suzhou) Co., Ltd.\\
  $^{3}$Shanghai Yunfan Intelligent Control Robot Technology Co., Ltd.\\
  $^{*}$Corresponding author: \href{mailto:shuang.liu@orillusion.com}{\texttt{shuang.liu@orillusion.com}}\par}
  \vspace{0.45em}
  {\small arXiv preprint -- manuscript version 3.6 -- 17 October 2026\par}
\end{center}

\vspace{0.25em}
\begin{abstract}
Language models acquired agents, harnesses and reusable skills only after data-rich pretraining had produced a strong base model. Embodied intelligence has no comparable corpus, so current routes to new manipulation capability either collect demonstrations for every task or keep a frontier model in the control loop for every episode. We study a third route: a strong general model acts only as an \emph{offline author} that writes, executes, diagnoses and repairs a manipulation skill in simulation until it passes a qualification exam on layouts drawn only after freezing; the frozen skill is then executed by code alone or by an inexpensive caller that reads nothing but its description, whereas concurrent self-evolving systems still run a frontier model in every episode at execution time. We call this skill-level recursive self-improvement \method and specify a compile protocol, with an isolated and audited author, hidden validation layouts that are never fed back, no-feedback and inherit-only control lineages and a tamper-evident ledger, and a reuse protocol that runs the identical frozen files with executors from another model family. On \rnTasks{} RoboTwin~2.0 tasks with cluttered tables, \rnTasksFrozen{} tasks were compiled into \rnFrozen{} frozen skills by Claude Code; \rnFrozenZeroRepair{} of the \rnFrozen{} came from a first draft that already met the targets and needed only qualification, at \$\rFrozenUSDMin--\rFrozenUSDMax{} of author cost each (\$\rProgramUSDPerFrozen{} per frozen skill once the \rnFailedCampaignsClaude{} campaigns that did not compile are counted). The other three tasks did not compile in \rnHardAll{} campaigns, including five with GPT-6 and a pre-registered experiment with 150 development layouts: on the hardest task every campaign scored \rSBThreeAllExamMin--\rSBThreeAllExamMax{} of 100 but had \rSBThreeAllAbortsMin--\rSBThreeAllAbortsMax{} safety aborts, so the pre-registered gate of 95\% success with zero aborts was not met, and a 100-episode exam bounds an abort rate only to \rAbortUpperZero\%. On hidden layouts, a repair session that inherits the previous version and its evidence delivered a better version than a from-scratch rewrite in \rHOCampPos{} of \rHOCampaigns{} controlled campaigns (sign test $p = \rHOSignP$; the selection-matched best-of-$k$ comparison is \rHOBestRepair--\rHOBestControl), and an inherit-only lineage, which revises the previous version with its own practice runs but without the runtime's development-layout feedback, shows that most of that gap comes from inheriting the previous version, with the development feedback adding a little more (\rFEightRICampPos{} of \rFEightN{} campaigns in the same direction, $p = \rFEightRISignP$). Six write-once campaigns found no detectable first-draft difference from practice, within a design that cannot exclude effects below about \rWOMDE{} points. The frozen skills measured for reuse retained \rPoolF{} successes on new layouts with no model and \rPoolO{} with a DeepSeek-flash caller at \$\rOdsCostMin--\rOdsCostMax{} of model cost per episode; no direct-control baseline was run, so no break-even is reported.
\end{abstract}


\vspace{0.4em}

\begin{figure}[!tp]
\centering
\resizebox{\textwidth}{!}{\input{figures/overview.tex}}
\caption{\textbf{Compile once, run cheap.} \protoA lets a strong author generate, repair and qualify a skill offline; an independent gate freezes it as an immutable \fs. \protoB runs the same files with executors that never see the skill code, on new layouts. Both protocols share one non-bypassable runtime.}
\label{fig:overview}
\end{figure}

\section{Introduction}

The capabilities that make language models useful as agents were built in a fixed order. Web-scale pretraining produced a strong base model first; agent loops, tool harnesses, reusable skills and protocols for calling them came afterwards, on top of that model. Embodied intelligence is trying to repeat the first step and has no corpus with which to do it. Robot foundation policies consolidate demonstrations, but each new task still asks for new demonstrations or post-training~\cite{kim2024openvla,pi07,amin2025pistar}. A second route places a frontier multimodal model inside a structured harness and lets it reason per episode; it solves tasks zero-shot but keeps the expensive model in the loop for every execution~\cite{chen2026showharness,tsui2026faea,guo2026robodawn,mosi2026roboicl}. A third route lets a frozen model evolve skills, critics or code through execution and revision~\cite{noematrix2026roborsi,ding2026zetta,wang2026shaper,lu2026aspire,zhang2026playful,bai2026practice}.

The third route produces a loop; it does not yet say what kind of object the loop produces or how its claims should be tested. Prior and concurrent systems have begun to evaluate frozen experience on scenes the loop did not see~\cite{li2026embodiedevo,song2026embodiedrsi}, but at execution time each episode still runs a frontier model, at a few dollars per episode where the cost is reported~\cite{li2026embodiedevo}, and neither of these systems separates learning from feedback from selection among attempts with a control. Three questions follow. First, can the product of the loop run without the model that wrote it, and with what loss when a cheaper model from another family calls it? Second, how much of the improvement across rounds is learning from feedback, how much is inheriting a previous version, and how much is trying several times and keeping the best? Third, what does the skill cost, counted as the full ledger of authoring, failed candidates, simulation and deployment, so that the claim that a skill is cheap has a denominator?

We take the position that the order of the language-model history can be inverted for robots: skills can come before the strong embodied base model, be compiled by a strong general model that already exists, and let robots start working, and generating structured data, in specific scenes now. This is a hypothesis, not a result. Its necessary precondition is empirical: a skill must be compiled cheaply enough, qualify on unseen layouts, run without the author, and cost little per episode. This paper tests that precondition.

\method separates authoring from execution (\cref{fig:overview}). In \protoA a strong model, whose weights never change, acts as an offline author: it drafts a skill package from a task card, runs it on fixed development layouts, reads the execution evidence, and revises, until the package meets a target on those layouts and on hidden validation layouts whose results it never sees. A qualification exam on layouts drawn only after freezing decides whether the package becomes a \fs, an immutable bundle of a skill description, restricted-Python skill code, a perception module and a manifest with hashes, seeds, environment fingerprint and compile cost. In \protoB the same bundle is executed by a code-only runner and by a caller from another model family that reads only the skill description, on layouts disjoint from every compile-time seed. We ask:

\begin{description}[leftmargin=1.7em,labelindent=0pt,itemsep=1pt]
  \item[Q1] How many repair rounds and how much author cost does it take to compile a task to a $95\%$ qualification bar with zero unsafe motions, and what stops the tasks that do not compile?
  \item[Q2] Is the improvement across rounds learning from feedback, or selection among independent attempts; within a repair round, how much comes from inheriting the previous version and how much from reading its execution evidence; and does the author's own testing within a session contribute?
  \item[Q3] When the frozen skill is executed by code alone or by a low-cost caller from another model family, how much success is retained?
  \item[Q4] What does a frozen skill cost per episode, with compilation amortized over deployments?
\end{description}

Our contributions are: (i)~an author--executor separation in which the unit of adaptation is a verified, portable \fs that executes with no model at all or with a cheap caller from another family, with two protocols that operationalize compile-once-run-cheap; (ii)~a credible measurement of what self-improvement contributes: an isolated and audited author, hidden validation layouts, exam layouts drawn after freezing and run to completion, no-feedback and inherit-only control lineages re-run on hidden layouts under a pre-registered design, and a per-decision cost ledger that includes every failed campaign; and (iii)~measurements on RoboTwin~2.0 with two author products that answer Q1--Q4, including three findings we did not anticipate: the frozen skills came mostly from first drafts, so the protocol's measured contribution on the tasks that compiled is qualification and freezing; the gate binds on rare safety aborts rather than on task completion, and more development layouts did not change that; and most of the per-round advantage of repair over rewriting comes from inheriting the previous version rather than from the runtime's development-layout feedback on it, while in-session self-testing shows no detectable effect on first drafts. We use the name \method because the author recursively refines a persistent external skill from that skill's own execution evidence; \cref{sec:terminology} states precisely what is and is not claimed under that name.

\section{Related Work}

\subsection{Improvement in weight space}
OpenVLA and later generalist policies map vision and language to actions through learned parameters~\cite{kim2024openvla}; $\pi_{0.7}$ extends this view with steerable, context-conditioned behavior~\cite{pi07}. Self-improvement in this family updates the policy from experience: RoboCat and self-improving embodied foundation models practice autonomously~\cite{bousmalis2023robocat,ghasemipour2025selfimproving}; SOAR lets foundation models propose and judge practice~\cite{zhou2024soar}; PLD probes a VLA's failure regions with residual RL and distills recovery data back~\cite{xiao2026pld}; RISE trains on imagined rollouts from a compositional world model~\cite{yang2026rise}; RECAP turns deployment experience into advantage-conditioned updates of $\pi^{*}_{0.6}$~\cite{amin2025pistar}; ENPIRE lets coding agents rewrite the training code itself on real hardware~\cite{xiao2026enpire}. World-action models couple prediction and action for contact-rich control~\cite{wang2026wam,liu2026oawam}. These routes need demonstrations, on-robot experience or training compute for every task. \method holds all weights fixed and evolves an external artifact; the comparison is weight-space adaptation versus artifact-space adaptation, not learning versus no learning.

\subsection{Frozen models in harnesses}
Code as Policies, Inner Monologue, VoxPoser and ReKep showed that language models can write executable control programs and use feedback~\cite{liang2022code,huang2022inner,huang2023voxposer,huang2024rekep}; REFLECT summarizes experience for failure explanation~\cite{liu2023reflect}. Recent harnesses put frozen models or policies behind structured observation--action interfaces: Show-Harness exposes discrete semantic actions to a VLM~\cite{chen2026showharness}, Harness~VLA orchestrates frozen VLA primitives with memory~\cite{zhang2026harnessvla}, Guava and hierarchical VLA agents study which harness components matter~\cite{liu2026guava,hu2026orchestrating}, AgentSpec studies scaffold composition~\cite{chen2026agentspec}, and EmbodiedSkills wraps VLA agents in skills defined as executable contracts with pre- and post-checks~\cite{wang2026embodiedskills}. FAEA applies a software-engineering agent to manipulation without demonstrations~\cite{tsui2026faea}, RoboDawn transfers VLM competence to RoboTwin~2.0 control zero-shot~\cite{guo2026robodawn}, and RoboICL adapts a frontier model to bimanual tasks from a few in-context demonstrations~\cite{mosi2026roboicl}. The clearest statement of the per-episode route is the early evaluation of GPT-6 Astra as a policy: given RGB views, robot state and a task recipe, with no depth and no privileged pose, the model chooses end-effector targets turn by turn and reaches 22.5\% success over 2{,}100 trials on the 42 RoboDojo tasks, far ahead of two other language models used the same way~\cite{zhang2026astra}. In all of these the model reasons during execution. Our runtime design follows theirs; our question is whether the reasoning can be done once, offline, and its result handed to something cheaper.

\subsection{Self-evolving skills}
Voyager grew a code skill library in an open-ended digital world~\cite{wang2023voyager}. Embodied systems now close the execute--diagnose--revise loop around frozen models (\cref{tab:related}). RoboRSI coordinates Manager, Planner, Engineer and Reviewer agents over a skill tree and consolidates stable tool-call chains into code~\cite{noematrix2026roborsi}; Zetta evolves runtime critics and recovery skills around a frozen policy~\cite{ding2026zetta}; SHAPER evolves skills and context-code harnesses with one frozen model as planner and optimizer~\cite{wang2026shaper}; ASPIRE refines code-as-policy programs into a library~\cite{lu2026aspire}; Playful Agentic Robot Learning acquires skills through self-proposed play~\cite{zhang2026playful}; EmbodiSkill adds skill-aware reflection~\cite{ju2026embodiskill}; VASO targets formally verifiable skills~\cite{yang2026vaso}; visual verification steers policies at inference time~\cite{zhang2026veritas}; Practice trains a skill learner that maintains a library for frozen executors~\cite{bai2026practice}; ELITE accumulates procedural strategies for a frozen VLM and retrieves them by procedural similarity~\cite{wei2026elite}. SkillSmith and SkCC compile software-agent skills into typed runtime interfaces~\cite{xu2026skillsmith,ouyang2026skcc}. Two concurrent systems are closest to ours. EmbodiedEvo lets a coding agent evolve reusable Skills and Lessons on one playground scene of each RoboDojo task, freezes them, and evaluates on new scenes, raising a matched-test success rate from 8.50\% to 30.17\% and transferring the experience to two other models (+8.86 and +5.50 points); at test time a GPT-6 Astra agent still reasons in every episode with that experience as context, with up to 100 accepted code executions per episode (29.0 on average with Skills and Lessons), at a mean model-API cost of \$5.73 per test episode, excluding the playground learning phase~\cite{li2026embodiedevo}. EmbodiedRSI is an agentic harness that adapts around the robot model: it co-evolves code and skills, holds competing hypotheses in a graph, chooses the physical experiments that discriminate between them, and reports 77.0\% on RoboCasa365, 86.8\% on LIBERO-Pro and 71.3\% on a real robot~\cite{song2026embodiedrsi}. Both evaluate frozen experience beyond the scenes it was learned on; neither measures execution without a large model, and neither includes a control that separates learning from selection. Our difference from them is in those two measurements, not in the loop. The repair loop itself has antecedents outside robotics: Reflexion and Self-Refine revise an answer from verbal feedback~\cite{shinn2023reflexion,madaan2023selfrefine}, Eureka refines reward code from execution statistics in robot simulation and compares against sampling more candidates~\cite{ma2023eureka}, and two audits found that self-repair of code pays only when external execution feedback is strong and that models do not improve their own reasoning without it~\cite{olausson2024selfrepair,huang2024selfcorrect}; our write-once and inherit-only results (\cref{sec:insession,sec:q2}) replicate that pattern in the embodied setting. \method could wrap any of these systems as its author. Its requirements are on the output and the evidence: an immutable package with a manifest, a gate between discovery and release, execution of the identical package by an executor from another model family, and a control that separates learning from selection.

\begin{table}[!tp]
\centering
\caption{Positioning against concurrent self-evolving and harness systems. Entries summarize each work's stated focus as of October 2026; ``--'' means not a stated focus, not that the capability is absent. The three right-hand columns are this paper's evaluation requirements, so the table shows where our evidence differs in kind, not that the other systems are weaker; several of them evaluate on real robots or on more tasks than we do. ``Runtime model'' is what reasons during deployment.}
\label{tab:related}
\footnotesize
\setlength{\tabcolsep}{5pt}
\begin{tabularx}{\textwidth}{@{}>{\raggedright\arraybackslash}p{2.8cm}>{\raggedright\arraybackslash}p{0.25\textwidth}*{4}{>{\centering\arraybackslash}X}@{}}
\toprule
System & Evolving object & Runtime model & Cross-family frozen execution & Exam layouts drawn after freezing & Learning vs.\ selection control \\
\midrule
RoboRSI~\cite{noematrix2026roborsi} & skill tree; code consolidated from stable tool chains & agents remain in loop & -- & -- & -- \\
Zetta~\cite{ding2026zetta} & critics and recovery skills around a frozen policy & frozen VLA & -- & -- & -- \\
SHAPER~\cite{wang2026shaper} & skills and context-code harness & same frozen model & -- & -- & -- \\
ASPIRE~\cite{lu2026aspire} & code-as-policy library & author model & -- & -- & -- \\
Practice~\cite{bai2026practice} & skill library via a trained learner & frozen executors & several executors & -- & -- \\
FAEA~\cite{tsui2026faea} & per-task policy through test-time iteration & frontier model, per episode & -- & -- & -- \\
RoboDawn~\cite{guo2026robodawn} / RoboICL~\cite{mosi2026roboicl} & none (zero/few-shot control) & frontier model, per episode & -- & -- & -- \\
Voyager~\cite{wang2023voyager} & code skill library (digital world) & same frozen model & -- & -- & -- \\
EmbodiedSkills~\cite{wang2026embodiedskills} & skills as executable contracts with pre- and post-checks & VLA agent & -- & -- & -- \\
ELITE~\cite{wei2026elite} & procedural strategies retrieved by similarity & frozen VLM, per episode & -- & -- & -- \\
EmbodiedEvo~\cite{li2026embodiedevo} (concurrent) & Skills and Lessons evolved on one scene per task, frozen for new scenes & GPT-6 Astra agent, per episode & experience read by two other models, still per episode & new scenes & -- \\
EmbodiedRSI~\cite{song2026embodiedrsi} (concurrent) & code and skills co-evolved, guided by competing hypotheses & robot model with an evolved harness & -- & evaluation seeds disjoint from evolution & -- \\
PhysicalRSI~\cite{physicalrsi2026} (concurrent, preview) & harness, skills and memory under propose--evaluate--select--inherit & API agent with policies and code skills & -- & -- & -- \\
\textbf{\method (ours)} & \textbf{\fs (immutable package)} & \textbf{none, or a cheap caller} & \textbf{measured (code; one DeepSeek caller; 3 skills)} & \textbf{yes} & \textbf{yes, on hidden layouts (8 + 3 campaigns)} \\
\bottomrule
\end{tabularx}
\end{table}

\subsection{Recursive self-improvement as a term}
The term goes back to Good's ultraintelligent machine~\cite{good1965speculations} and Schmidhuber's G\"odel machine, which rewrites its own code under proof~\cite{schmidhuber2007godel}; STOP and the Darwin G\"odel Machine brought it to frozen language models that improve an external scaffold or agent code against held-out tasks~\cite{zelikman2023stop,zhang2025dgm}. Duan et al.\ order recursive self-improvement into five levels, from executing improvements to improving the improvement process itself, and name embodied intelligence as an application~\cite{duan2026lastai}. Surveys of self-evolving embodied AI and post-deployment humanoid self-improvement describe memory, task, environment, embodiment and model self-evolution with a slow outer loop~\cite{feng2026selfevolving,nguyen2026humanoids}. In software agents the term is now used for loops that improve an external artifact while the model stays frozen: RRSI regularizes recursive improvement of prompts, control flow, tools and memory so that reusable mechanisms win over benchmark-specific ones~\cite{xia2026rrsi}; AIDE$^{2}$ lets an outer agent rewrite the harness of an inner research agent and keeps changes that survive held-out benchmarks~\cite{srikanth2026aide2}; MetaSkill-Evolve evolves task skills on a fast loop and a meta-skill on a slow one~\cite{wang2026metaskill}; RSIAgent builds environment memory by exploration without parameter updates~\cite{zhu2026rsiagent}. In embodied systems the term is now used concurrently by several groups: EmbodiedRSI for hypothesis-guided co-evolution of code and skills in a harness around the robot model~\cite{song2026embodiedrsi}, and PhysicalRSI, a preview release without a paper, for a propose--evaluate--select--inherit loop over a harness, skills and memory~\cite{physicalrsi2026}. Our loop uses the term in the artifact-level sense, sits at the first three of Duan et al.'s levels, does not claim the fifth, and calls a gain self-improvement only where a control lineage shows what the gain is relative to; \cref{sec:terminology} makes the boundary explicit.

\subsection{Hidden physical properties and touch}
Text corpora contain almost no quantitative tactile knowledge, and touch data are scarce; Sparsh learns touch representations from tactile sensors~\cite{higuera2024sparsh}, TacGen synthesizes tactile latents from RGB to mitigate the scarcity~\cite{ye2026tacgen}, and DeliGrasp asks an LLM to infer mass, friction and compliance from object semantics to set grasp forces~\cite{xie2024deligrasp}. These works complete the missing modality. Our route is complementary: a skill can encode a strategy that avoids needing the modality, such as supporting a fragile object from below rather than estimating a safe grip force; \cref{sec:discussion} discusses where that substitution stops working.

\subsection{Evaluation validity}
LIBERO measures lifelong transfer~\cite{liu2023libero}; LIBERO-PRO and LIBERO-Plus expose memorization and decompose robustness into perturbation dimensions~\cite{zhou2025liberopro,fei2025liberoplus}; RoboTwin~2.0 provides domain-randomized bimanual tasks and a public expert generator~\cite{chen2025robotwin2}; VLABench, ESI-Bench and BEHAVIOR-1K extend scope~\cite{zhang2024vlabench,hong2026esibench,li2024behavior}. Recent audits find shortcuts, repeated test exposure and low trial counts behind reported progress~\cite{jiang2026benchmarking}, and cost-controlled evaluation of agents asks that accuracy be reported with the dollars spent to reach it~\cite{kapoor2024agents}. These concerns apply with particular force to self-evolving systems, whose tasks are also their discovery set, and to agentic authors with shell access, which, we found, read other campaigns' work when nothing stops them (\cref{sec:isolation}). We respond with author isolation and session audits, hidden validation layouts, exams drawn after freezing, run-to-completion exams, seeds disjoint across protocols, and a full ledger of failed candidates.

\section{Setting and Terminology}
\label{sec:setting}

\subsection{Components}
A task instance is a partially observed process in which a dual-arm robot receives observations, executes primitive actions and transitions under simulator dynamics. Skills do not emit joint torques. A fixed \emph{kernel} exposes six primitives (\cref{tab:api}, \cref{app:package}): \texttt{observe}, \texttt{ee\_pose}, \texttt{gripper\_state}, \texttt{move\_pose}, \texttt{gripper} and \texttt{wait}, and translates them into planned trajectories under workspace, joint and drop monitors with safety abort. The system has an \emph{author} $A$ (a strong model product used only offline), an \emph{executor} $E$ (a code-only runner or a caller model), the \emph{skill package} $K$, the kernel $D$ and an \emph{evaluator} that reads privileged simulator state to decide success. The author never reads the evaluator; the executor never reads the skill code.

\subsection{Compile--reuse boundary}
We call the setting \emph{target-task weight-update-free}: neither author nor executor parameters change for the target task; no new human teleoperation or expert trajectories are collected for it; simulation rollouts, environment predicates, structured traces and the declared primitive API may be used; the external skill may change during compilation and never afterwards. Foundation-model pretraining is not zero, simulator interactions are data and success predicates are supervision, so we avoid the phrases ``zero data'' and ``from scratch''. The boundary between \protoA and \protoB is a hash: the three skill files and the manifest are content-addressed at freezing, and every reuse episode verifies them before execution.

\subsection{What we call RSI, and what we do not}
\label{sec:terminology}
Ordinary episode retry is not self-improvement, and a higher final score alone does not show that a process improved itself. We use four stages that match the released code. \textbf{R0 generate}: independent drafts of the package from the task card. \textbf{R1 repair}: recursive refinement of a persistent package, in which version $k{+}1$ is written by the author from version $k$ and from the execution evidence of version $k$ on the development layouts. \textbf{R2 qualify}: an exam on layouts drawn after the package is frozen. \textbf{R3 reuse}: execution of the frozen package by executors that did not author it. The recursion is over the artifact, not over the author: the author's weights, prompts, tools and diagnostic procedure are fixed for the whole run, as in recent harness-level recursive self-improvement for software agents~\cite{xia2026rrsi,srikanth2026aide2}. In the five-level ordering of Duan et al.~\cite{duan2026lastai}, R1 exercises improvement execution, strategy choice and experience acquisition, but not meta-level improvement of the improver, which we leave to future work.

Two kinds of self-improvement occur in a campaign and we measure them separately. \emph{In-session} improvement happens inside one author session: the product writes, tests on practice layouts or in its own simulator account, and revises before delivering; every R0 draft already contains it. \emph{Cross-session} improvement is R1: each round is a new session that inherits the package and the evidence of all earlier versions. We define the \emph{RSI delta} of a campaign as exam success minus the hidden-validation baseline of the selected R0 draft, which is the cross-session part; the in-session part is the difference between that baseline and the baseline of a \emph{write-once} draft produced under a condition in which the author cannot run the package at all (\cref{sec:ablations}). A repair round contains three separable ingredients: \emph{selection} (the runtime keeps the best version), \emph{inheritance} (the author starts from the previous version instead of from the task card) and \emph{development feedback} (the author reads the runtime's per-layout evaluation of the previous version on the development block). All lineages keep the author's own practice runs on practice layouts, so what the control lineages remove is not testing but the runtime's evaluation. They remove the ingredients one at a time under the same selection rule: a no-feedback rewrite has neither inheritance nor development feedback, an inherit-only lineage has inheritance without development feedback, and repair has both; the differences repair$-$inherit, inherit$-$control and repair$-$control therefore measure the marginal effect of the development feedback, the effect of inheritance, and their sum. Claims about ``self-improvement'' in this paper are claims about R1 and are licensed only by those controls, evaluated on layouts that none of the lineages has seen.

\subsection{The \fs package}
\label{sec:package}
A package has four files. \texttt{SKILL.md} (at most 20\,KB) starts with a front matter naming the skill and its atoms and then documents each atom, a standard flow and a recovery section; it is the only file a caller may read. \texttt{skills.py} defines \texttt{ATOMS}, each atom as \texttt{atom(api, args) -> \{ok, ...\}}, and \texttt{fallback(api)}, a complete code-only policy. \texttt{perception.py} defines \texttt{detect(rgb, depth, calibration)}, which must return, from the RGB-D image, exactly one component per task role (a pixel box, a world position, a point count and any further numeric fields the author chooses) plus a table-height estimate; a frozen CLIP model, fixed by the runtime, scores the author's crop boxes against the role names and assigns the roles, and \texttt{observe()} hands the resulting components to the skill. Perception is therefore the author's, role naming is the runtime's. \texttt{rationale.md} records design decisions and is excluded from the frozen bundle. Both Python files are \emph{restricted Python}: no imports, no \texttt{while}, \texttt{try}, classes or lambdas, a builtin allowlist, and no underscore names; a static validator rejects anything else, and the runtime enforces that skills touch the simulator only through the API. The frozen bundle adds \texttt{MANIFEST.json}: file hashes, runtime version (the API contract), environment fingerprint, gate outcome, the development, validation, exam and caller-check seeds, and the compile-cost summary (\cref{app:package}).

A package runs in two modes. In \textbf{fallback} mode \texttt{fallback(api)} runs alone; no model is involved. In \textbf{orchestrated} mode a caller reads \texttt{SKILL.md}, receives the task, one observation and the execution log so far, and returns a plan of at most eight steps, each an atom call, an observation or a state read; the runtime executes the plan, stops at the first \texttt{ok:\,false}, and asks again, at most four times per episode during compilation. The caller has no tools and cannot move the robot except through atoms. Per-episode limits are 90 primitive calls, 90 simulated seconds and 240 wall-clock seconds excluding caller latency. After the skill returns, the robot holds still for two simulated seconds before the official success predicate is read, so that an object released at the last moment is judged where it comes to rest.

\section{\protoA: Compiling a Skill}
\label{sec:protoA}

\protoA is a scripted coordinator that never calls a model for judgement and never interprets results for the author. Its stages are summarized in \cref{alg:compile}; the parameters are in \cref{tab:params} (\cref{app:gate}).

\paragraph{Preparation and lock.}
A task configuration names the instruction and the object roles; everything else is generated from the official RoboTwin~2.0 source. The task card given to the author contains the instruction, the roles, the success predicate source, object generation parameters, one example observation and the robot's gripper calibration; the expert generator is on a deny list and is never shown. All campaigns use one scene tier, \emph{clutter}: the official \texttt{demo\_clean} configuration with up to ten task-irrelevant distractor objects on the table (none in 2\% of layouts), fixed lighting and background, and no table-height randomization. A phase-0 health check materializes 30 layouts, runs the official expert ten times to confirm feasibility and to fix the speed bar, checks reachability, and calibrates the gripper; it stops the campaign on red lights (materialization failure above 10\%, expert success below 30\%, an unreachable role, an inseparable gripper reading). The gate specification (success bar, safety, speed factor, regression rule) is written before any author session and never edited. Three seed blocks are then materialized: 50 \emph{development} layouts (seeds 70000--70049), 50 \emph{validation} layouts (71000--71049) whose results are never shown to the author, and a practice range (68000--68999). A layout that the official scene generator fails to stabilize is dropped for every version, so the usable blocks had \rDevNMin--\rDevNMax{} development and \rValNMin--\rValNMax{} validation layouts per task; targets are stated as fractions of the usable block.

\paragraph{R0: independent drafts.}
Two author sessions start in separate working directories with identical inputs and write a first package each. Each package passes static validation and runs in fallback mode on the development layouts. The better draft (more successes; ties broken by fewer primitive calls) becomes the lineage head; its directory, including the author's own notes, is copied into the repair workspace. Before any repair, the head is run once on the validation layouts; this \emph{validation baseline} is the ``before'' measurement for the whole campaign, because the development score of a selected draft is biased upward by the selection itself.

\paragraph{R1: repair until target.}
Each round starts a fresh author session of at most one hour in the repair workspace. The author sees every earlier version, its per-layout outcomes on the development layouts (success, failure class, call counts, safety events, and for failed layouts the primitive trace and observation fields), the verdict that compared it with the head, the current head, and its own notes. It must deliver a complete package; the runtime runs it in fallback mode on the development layouts and replaces the head only if the new version has strictly more successes. When the head reaches the development target ($\geq 96\%$ of usable layouts), it is run once on the validation layouts; if it reaches the validation target ($\geq 94\%$) R1 ends, otherwise the author is told only that ``the package did not yet meet the acceptance bar'' and repair continues. R1 also ends after ten rounds, after three consecutive rounds without improvement, or when the author's observable cost reaches \$150 or its session time ten hours; in every case the campaign proceeds to R2, so that each campaign ends with an exam.

\paragraph{Practice and the author's own experiments.}
The author may run up to 30 practice episodes per round on practice seeds through a gateway, and, since runtime 5.5.0, holds an unprivileged account on the simulation host where it can run its own experiments with the installed simulator on a shared GPU slot. These are the means of in-session improvement; both are logged, and any episode not scheduled by the coordinator is listed as unscheduled.

\paragraph{Control lineages.}
\label{sec:control}
The replacement rule alone raises the head's development score over rounds even if the author learns nothing, because each round is another draw on the same layouts. To separate learning from selection, when repair round $k$ starts the coordinator also starts control draft $C.k$: a new working directory with exactly the R0 inputs and prompt, no history and no feedback, evaluated in fallback mode on the same development layouts. Draft and control directories carry random names. Two comparisons follow. The \emph{best-of-$k$} comparison, with $r_j$ and $c_j$ the fallback successes of repair $j$ and control $j$ on the development layouts and $R_0$ the set of R0 scores, is
\begin{equation}
\mathrm{Learn}(k) = \max(R_0 \cup \{r_1..r_k\}) - \max(R_0 \cup \{c_1..c_k\}),
\label{eq:learn}
\end{equation}
which matches attempts and selection rule on both sides. The \emph{per-round} comparison asks whether the version delivered by repair round $k$ is better than the version delivered by control draft $k$; because the repair session has seen the development layouts and the control has not, this comparison is made on the hidden validation layouts, on which every $R1.k$ and $C.k$ is run once after the campaign ends (runtime 5.8.0; report-only, never fed back). The two comparisons are not equivalent: repair starts from the better of two drafts and keeps only strict improvements, whereas each control is a single fresh draw, so the per-round comparison also credits repair for not discarding an already selected starting point; we therefore also report each lineage relative to the campaign's own starting point. The control was enabled on a subset of campaigns because it costs one author session per repair round.

A second experiment, pre-registered on 2026-10-10 before any of its campaigns ran (runtime 5.9.0; registration text and deviation record in \cref{app:prereg}), adds a third lineage to separate inheritance from feedback. An \emph{inherit-only} lineage $I.k$ is copied from the R0 starting workspace with the author's notes and practice records kept and the starting version's feedback and verdict deleted; each round the author revises the previous version with the same 30 practice episodes per round, whose results it sees, but receives none of the runtime's development-layout evaluation (no per-layout outcomes, traces or verdicts of any version), only whether the runtime replaced the head. Rounds are matched to the repair lineage, the inherit-only author has its own host account, its cost is not charged against the repair caps, and its versions are not examined. The same experiment raised the development block from 50 to 150 seeds (70000--70149; \rFEightDevNMin{} usable after materialization) on the task that had never compiled, to test whether a block large enough to show a 1--3\% abort rate lets repair clear the gate; the validation block, the targets, the round and patience limits and the gate were unchanged. Its primary outcome, fixed in advance, is the number of campaigns frozen out of three, reported descriptively; its secondary outcomes are the RSI delta and the abort count of each campaign, and the per-round held-out comparisons repair$-$inherit, inherit$-$control and repair$-$control with the campaign as the unit of the sign test, which with three campaigns has a minimum attainable one-sided $p$ of 0.125. The registration states that the two changes, more layouts and the new lineage, were introduced together, so a change in the primary outcome could not be attributed to either alone.

\paragraph{R2: qualification exam.}
Before the exam the coordinator verifies the hash-chained ledger, the runtime manifest and the official code tree; any change voids the campaign. The head is frozen; only then are 100 exam seeds drawn uniformly from $[10^5, 10^6)$, disjoint from all earlier seeds. The exam runs in fallback mode and passes if successes are at least 95 of 100, no \emph{safety abort} occurs (an object dropped below the table plane, an end effector leaving the workspace box, a joint beyond its limit, or an invalid pose request; distractors are not monitored), the median simulated time of successful episodes is at most twice the expert median from phase~0, and the frozen version is not worse on the development layouts than any version previously accepted as head. In every episode of both protocols the official success predicate is evaluated only after a 2\,s settle in which the robot holds still while safety monitoring continues (not counted against the time budget; the phase-0 expert is judged the same way), so that an object released just before the check cannot be judged while still moving. Failure is decided as soon as more than five episodes have failed, but the exam always runs to completion so that failed campaigns report a comparable success rate; the outcome is \texttt{frozen}, \texttt{failed exam} (R1 reached its targets but the exam failed) or \texttt{not compiled} (R1 ended on a budget rule). No retake is allowed. The rule is a point-estimate threshold: a skill that passes with 95 of 100 has a Wilson lower bound of only \rWilsonLBNinetyFive\%, and $n{=}100$ was chosen so that the interval at the threshold has a half-width of about \rWilsonHalfWidthNinetyFive{} points, not so that the bar is demonstrated with confidence; we report the interval beside every verdict. A passing package is written as a \fs with its manifest.

\paragraph{Caller check.}
After freezing, the coordinator runs 20 episodes on fresh seeds in orchestrated mode with the author product as caller, reading only \texttt{SKILL.md}. The result does not affect the gate; it is recorded so that a description that code can execute but callers cannot follow is detected before \protoB, with a warning if it falls more than 15 points below the exam score.

\paragraph{Author isolation and audit.}
\label{sec:isolation}
The author runs with a shell, network access and a simulator account because practice needs them. In an earlier version of the protocol without isolation, an audit of 58 campaigns' session transcripts found that in five the author had read and copied skill code, notes or scores of other campaigns on the same machine, and that one formal campaign's starting draft was 0.96 similar to an earlier candidate for the same task and had been tested on hidden validation layouts; none of those campaigns is reported here. Since runtime 5.5.0 the author process runs in an OS sandbox restricted to its own working and home directories, each working directory has its own unprivileged user and key on the simulation host with a read-only view of its permitted files, validation and exam layouts and other campaigns are unreadable, and every campaign ends with an automatic audit that pairs each tool call in the session transcripts with the result returned to the model. The audit grades a campaign \emph{clean}, \emph{metadata} (another campaign's identifier seen in a directory or process listing, no content), \emph{review} (a path of another campaign appeared in a result; checked by hand, no content read), \emph{contaminated} (content of another campaign read) or \emph{unaudited} (transcripts missing); the last two invalidate the campaign.

\paragraph{Tamper evidence.}
The results ledger is hash-chained, the runtime is pinned by a manifest recorded at campaign creation, the official code tree is hashed, exam seeds do not exist until freezing, and feedback generation asserts that no validation or exam seed appears in what the author sees. Each campaign records an environment fingerprint (GPU, driver, PyTorch, planner build); campaigns with different fingerprints are never compared, because planner numerics change outcomes on identical layouts.

\paragraph{Ablations and post-hoc evaluations.}
\label{sec:ablations}
Two ablation conditions stop after the validation baseline: \emph{practice only} removes the simulator account (runtime 5.6.0), and \emph{write once} removes practice as well, telling the author that the package cannot be run before delivery (5.7.0); the task card and the rest of the prompt are otherwise identical. Both conditions were added after the first formal campaigns had run and their first-draft scores were known; they were not pre-registered. Two report-only commands were added after the first formal results and are not pre-registered: a \emph{re-evaluation} of the exam version on 300 further layouts drawn disjointly from all earlier draws (5.5.3), and the \emph{held-out} re-run of every repair and control version on the validation layouts described above (5.8.0). Neither changes any campaign outcome.

\begin{algorithm}[t]
\caption{\protoA: compile one task with one author}
\label{alg:compile}
\begin{algorithmic}[1]
\Require task card $T$, author $A$, dev seeds $S_d$, validation seeds $S_v$, gate $G$ (locked)
\State $K_a, K_b \gets A(T)$ in separate directories \Comment{R0}
\State head $\gets \arg\max_{K\in\{K_a,K_b\}} \mathrm{fallback}(K, S_d)$
\State $b_v \gets \mathrm{fallback}(\text{head}, S_v)$ \Comment{validation baseline, hidden}
\For{$k = 1, 2, \dots$ until a stop rule fires} \Comment{R1}
  \State start control draft $C.k \gets A(T)$ with no history \Comment{optional}
  \State start inherit-only draft $I.k \gets A(I.{k-1})$ without development feedback \Comment{second experiment}
  \State $K_k \gets A(\text{head}, \text{evidence}(\cdot, S_d), \text{history})$
  \If{$\mathrm{fallback}(K_k, S_d) > \mathrm{fallback}(\text{head}, S_d)$} head $\gets K_k$ \EndIf
  \If{$\mathrm{fallback}(\text{head}, S_d) \ge 0.96|S_d|$ \textbf{and} $\mathrm{fallback}(\text{head}, S_v) \ge 0.94|S_v|$} \textbf{break} \EndIf
\EndFor
\State verify ledger chain, runtime manifest, official tree; freeze head \Comment{R2}
\State $S_e \gets$ 100 seeds drawn now, disjoint from all prior seeds
\State outcome $\gets G(\mathrm{fallback}(\text{head}, S_e), \text{aborts}, \text{speed}, \text{regr.})$
\If{outcome = passed} write \fs with manifest; run caller check \EndIf
\State audit session transcripts; \Return outcome, compile cost, curves
\end{algorithmic}
\end{algorithm}

\section{\protoB: Measuring Reuse}
\label{sec:protoB}

\protoB takes a \fs and never modifies it. The version run here is the minimal one implemented inside the compile runtime (5.8.0): it verifies the three file hashes against the manifest, checks that the runtime version and environment fingerprint match, draws 50 seeds from the exam range excluding every development, validation, exam and caller-check seed recorded in the manifest, and executes every arm on the same 50 layouts.

\paragraph{Executor arms.}
\textbf{F} runs \texttt{fallback(api)} with no model. \textbf{O-ds} is an orchestrated run with \texttt{deepseek-flash} as caller, chosen because it is inexpensive, fast, and from an organization different from the author's; the caller sees only \texttt{SKILL.md}, receives the identical prompt template used in the caller check, and may decide up to twelve times per episode (the mean number of decisions is reported). Decoding used temperature 0 with an output limit of 32{,}768 tokens including reasoning (hosted decoding at temperature 0 is not bit-reproducible). A plan that fails to parse is an invalid decision charged to the caller, not an infrastructure failure. We did not run callers from further model families, a strong-model caller, or the direct-online baselines of the full design (\cref{sec:limits}).

\paragraph{Ledger.}
Every decision records the arm, exact model identifier, UTC time, input, cached and output tokens, the price applicable at that time, cost, latency and the raw response; every episode records success, failure class, safety events, decisions, primitive calls, simulated and wall-clock time, and atom entry and exit marks. Costs are provider-reported usage multiplied by the price captured on 2026-10-08 (peak or off-peak by UTC hour, \cref{app:prices}); unavailable usage is recorded as null, never estimated.

\paragraph{Statistics.}
Success rates carry Wilson 95\% intervals~\cite{wilson1927probable}; safety-abort rates carry Clopper--Pearson exact intervals. Counts pooled over skills are sums and are labelled as such. Arms share seeds, so paired tests are possible in principle; with the outcomes observed here the number of discordant pairs is too small for any test to separate the arms, and we report the counts. Per-episode cost with compilation amortized over $N$ deployments is $C/N + c_X$, with $C$ the author cost from the manifest and $c_X$ the arm's mean model cost per episode; simulation compute is listed but not priced.

\section{Experimental Setup}
\label{sec:setup}

\paragraph{Environment.}
All experiments use RoboTwin~2.0~\cite{chen2025robotwin2} with the \texttt{aloha-agilex} dual-arm embodiment, the clutter tier described above, and cuRobo-based planning behind the primitive API. The planner's collision world contains only a static table box ($0.7 \times 2 \times 0.04$\,m) and the robot's self-collision; task objects and distractors are not in it, so a planned arm trajectory avoids the table and the robot but not the objects on the table, and keeping clear of them is the skill's responsibility through the poses it requests. Everything ran on one \rGPU{} (driver \rDriver, PyTorch \rTorch, OptiX denoiser) with four concurrent simulator slots; the environment fingerprint was identical for every campaign and evaluation reported here. The compile runtime was frozen at version 5.5.0 before the first formal campaign and received seven increments during the study (5.5.1--5.8.0: sandbox and GPU-slot fixes, the report-only evaluations, audit support for the second product, quota-exhaustion handling, and the minimal reuse protocol), each dated in the released protocol; none changed the default author prompt or a gate rule; the primary Claude campaigns ran on \rRuntimeMin--\rRuntimeMax, the GPT-6 campaigns on \rGPTRuntimeMin--\rGPTRuntimeMax, \protoB on 5.8.0, and the pre-registered second experiment on 5.9.0, which added the inherit-only lineage and the larger development block and changed nothing else.

\paragraph{Tasks and campaigns.}
Seven RoboTwin~2.0 tasks were compiled (\cref{tab:compile}). They were chosen after piloting, not before. A pilot on the fully randomized scene tier found first drafts above 82\% on four tasks, so the clutter tier was used throughout; pilots on the clutter tier from 2026-09-28 to 09-30, run on an earlier runtime without author isolation, were used to pick tasks whose best first draft scored between 30\% and 80\% (\texttt{place\_phone\_stand} 31\%, \texttt{stack\_bowls\_two} 77\%, \texttt{move\_can\_pot} 73\%, and \texttt{stack\_bowls\_three}), with one deliberately harder task (\texttt{hanging\_mug}) and one deliberately easier task (\texttt{place\_burger\_fries}) added once each; on 2026-10-01 all formal campaigns were restarted on the isolated runtime 5.5.0, and \texttt{place\_dual\_shoes} was added after a further pilot on 10-06 (first drafts 82--84\%). In the formal campaigns the first drafts of the mid-difficulty tasks were already near the bar (\cref{tab:compile}), higher than their pilots, which we report as observed; the pilot runtime and the formal runtime differ in isolation and in the author's host account, and we do not attribute the difference to either. The second experiment chose \texttt{stack\_bowls\_three} because it had never compiled. Replication is unbalanced by design: tasks that did not compile were repeated to see whether more budget helps (\texttt{stack\_bowls\_three} six times, \texttt{place\_dual\_shoes} three), giving \rnClaude{} campaigns with the primary author; the no-feedback control was enabled on \rnControl{} of them, on four tasks, of which \rnControlComparable{} needed at least one repair round and therefore have a comparison. Five further campaigns with a second author product (\texttt{hanging\_mug}, and two each of \texttt{stack\_bowls\_three} and \texttt{place\_dual\_shoes}) test whether the pattern depends on the product. The pre-registered second experiment (\cref{sec:control}) ran three more campaigns on \texttt{stack\_bowls\_three} with 150 development seeds and all three lineages, after the 20 primary campaigns and with their outcomes known.

\paragraph{Authors and caller.}
The primary author was \rClaudeProduct{} running \texttt{\rClaudeModel} at high effort, 2026-10-01 to 10-09; the second author was \rGPTProduct{} running \texttt{\rGPTModel} at high reasoning, 2026-10-07 to 10-10. Both were used as shipped through their subscription channels, in a clean per-campaign home directory linked only to their credentials, with session environment variables removed and the locale fixed to English, so that the tested object is the product and not the operator's configuration. Author cost is the API-equivalent amount the product reports; the second product does not report one, and its cost is recorded as unobservable rather than estimated. The \protoB caller was \texttt{deepseek-flash} (DeepSeek-V4.1-Flash on the provider's price page), accessed through its OpenAI-compatible API on 2026-10-09.

\paragraph{Budgets.}
Per campaign: two R0 drafts, at most ten repair rounds, patience three, a \$150 author-cost cap and a ten-hour author-time cap; development target $\geq 96\%$ and validation target $\geq 94\%$ of usable layouts; exam $n{=}100$ run to completion; caller check $n{=}20$. \protoB: 50 seeds per frozen skill, both arms on the same seeds.

\paragraph{What is reported.}
For Q1: repair rounds, author cost and time, simulation episodes, wall-clock, exam success and safety aborts with intervals, outcome, and the RSI delta. For Q2: the write-once ablation, the per-round held-out comparison, $\mathrm{Learn}(k)$, and what repair sessions read. For Q3: per-arm success with intervals, safety aborts, decisions and invalid decisions. For Q4: cost per episode and amortized cost. In addition: the caller check, the isolation audit, speed, and the episode ledger as data.

\section{Results}
\label{sec:results}

\begin{table}[!tp]
\centering
\caption{Every compile campaign. R0 is the better first draft in fallback mode on the usable development layouts; val.\ base is the same draft on the hidden validation layouts before repair; curve is the development score of the head at the start and end of R1; cost is the author's reported API-equivalent cost (unobservable for the second product); hours are author session hours; wall is campaign wall-clock including all simulation; exam is fallback successes of 100 on layouts drawn after freezing, with Wilson 95\% interval; aborts are safety aborts in the exam; $\Delta$ is exam minus validation baseline in points; caller is the post-freeze caller check (author product, 20 layouts); audit is the isolation grade. Above the rule: \rClaudeProduct{} (\texttt{\rClaudeModel}); below: \rGPTProduct{} (\texttt{\rGPTModel}).}
\label{tab:compile}
\scriptsize\setlength{\tabcolsep}{3.2pt}
\begin{tabular}{@{}lcccccccccccccccc@{}}
\toprule
Task & Rep & Ctrl & R0 & Val.\ base & Rounds & Curve & Cost \$ & Hours & Wall h & Exam & 95\% CI & Aborts & $\Delta$ & Outcome & Caller & Audit \\
\midrule
\tabCompileRows
\bottomrule
\end{tabular}
\end{table}

\subsection{Q1: compiling to the qualification bar}
\label{sec:q1}

\Cref{tab:compile} lists every campaign. \rnFrozen{} of \rnClaude{} primary-author campaigns produced a \fs, on \rnTasksFrozen{} of \rnTasks{} tasks; \rnFailedExam{} reached the R1 targets and failed the exam, and \rnNotCompiled{} ended R1 on a budget rule. The seven frozen skills cost \$\rFrozenUSDMin--\rFrozenUSDMax{} of author budget (median \$\rFrozenUSDMedian), \rFrozenHoursMin--\rFrozenHoursMax{} author-hours in \rFrozenSessionsMin--\rFrozenSessionsMax{} sessions, and \rFrozenWallMin--\rFrozenWallMax{} wall-clock hours including every simulated episode. Five of the seven needed no repair round: the better of two first drafts met the development and validation targets, and the exam confirmed it. The two others needed one and two rounds. Campaigns that did not compile cost \$\rNCUSDMin--\rNCUSDMax{} (median \$\rNCUSDMedian), with \texttt{stack\_bowls\_three} alone consuming \$\rSBThreeUSDTotal{} over six campaigns; across all \rnClaude{} campaigns the author bill was \$\rAllUSDTotal{} for \rAllHoursTotal{} session-hours, and campaigns ran \rSimEpisodesMin--\rSimEpisodesMax{} simulated episodes (median \rSimEpisodesMedian). No campaign failed the speed or regression rule: exam versions ran at \rSpeedRatioMedian$\times$ the expert's median simulated time (at most \rSpeedRatioMax$\times$ against a limit of $2\times$).

Exam success itself was high: \rnRateBar{} of \rnClaude{} campaigns scored at least 95, including every campaign on five of the seven tasks; the eight below 95 are five of the six three-object-stack campaigns (\rExamsStackBowlsThree{} over the six) and the three bimanual shoe campaigns (\rExamsPlaceDualShoes). What separated passing from failing among the twelve was the safety rule, to which we return next. The RSI delta, exam minus validation baseline, is near zero for zero-repair campaigns, as it must be (the same version measured on two layout sets); on the \rnRepaired{} campaigns with at least one repair round it averaged \rDeltaMeanRepaired{} points (median \rDeltaMedianRepaired; range \rDeltaMinRepaired{} to \rDeltaMaxRepaired; positive in \rnDeltaPositive{} of \rnRepaired), with the largest gains (\rBigDeltaList) on the lowest starting points. Baseline and delta are strongly anticorrelated ($r = \rCorrBaselineDelta$), partly by construction; the more telling statistic is the exam itself: repaired campaigns that started below 80\% on validation averaged \rExamMeanBaseLow{} on the exam and those that started at or above 80\% averaged \rExamMeanBaseHigh. The baseline and the exam are nearly uncorrelated on the repaired campaigns ($r = \rCorrBaselineExam$): repair lifts different starting points to a ceiling that is set mostly by the task (\cref{fig:learning}b).

The pre-registered second experiment (\cref{tab:feight}) repeated \texttt{stack\_bowls\_three} three times with \rFEightDevNMin{} usable development layouts instead of 43. Its primary outcome is \rFEightFrozen{} frozen of \rFEightN: the repair lineages reached \rFJOneDevBest, \rFJTwoDevBest{} and \rFJThreeDevBest{} on development, one to three layouts short of the 96\% target, ended on the patience rule after \rFEightRoundsList{} rounds, and scored \rFEightExams{} on the exam with \rFEightAborts{} safety aborts. With the six earlier campaigns the task is \rSBThreeAllFrozen{} of \rSBThreeAllCampaigns{} frozen at \rSBThreeAllExamMin--\rSBThreeAllExamMax{} success and \rSBThreeAllAbortsMin--\rSBThreeAllAbortsMax{} aborts per 100. The RSI deltas were \rFEightDeltaList{} points, so two of the three campaigns moved a low starting point most of the way to the ceiling; the task's ceiling, not the author's progress toward it, is what stays below the gate. Each campaign cost \$\rFJOneUSD--\rFJThreeUSD{} of author budget across its three lineages (\$\rFEightCostRepair{} repair, \$\rFEightCostControl{} control and \$\rFEightCostInherit{} inherit-only in total) and \rFEightWallMin--\rFEightWallMax{} wall-clock hours.

\begin{table}[!tp]
\centering
\caption{Pre-registered second experiment (runtime 5.9.0): \texttt{stack\_bowls\_three} with 150 development seeds (\rFEightDevNMin{} usable) and three lineages. Dev best is the repair lineage's best development score; cost is the author cost of the repair / control / inherit-only lineages; wall is campaign wall-clock in hours; exam is fallback successes of 100 with Wilson interval; $\Delta$ is exam minus validation baseline. $^{\dagger}$j3's exam was redrawn after a voided round (\cref{app:prereg}).}
\label{tab:feight}
\scriptsize\setlength{\tabcolsep}{4pt}
\begin{tabular}{@{}lcccccccccll@{}}
\toprule
Run & Rounds & Dev best & Val.\ base & Cost \$ (R / C / I) & Wall h & Exam & 95\% CI & Aborts & $\Delta$ & Outcome & Audit \\
\midrule
\tabFEightRows
\bottomrule
\end{tabular}
\end{table}

\subsection{The gate binds on rare safety aborts, not on task completion}
\label{sec:safety}

Twelve primary-author campaigns scored at least 95 and seven passed; the five that did not were stopped by the safety rule alone, with \rSafetyOnlyAbortsList{} aborts in 100 episodes at success rates of \rSafetyOnlyExamMin--\rSafetyOnlyExamMax. Of the eight campaigns below 95, the five three-object-stack campaigns also aborted often (4--16 per 100) and the three shoe campaigns rarely (0--1). Every one of the \rExamAbortsTotal{} exam aborts across the 25 campaigns was a physical envelope violation: \rExamAbortsWorkspace{} were objects pushed out of the workspace box and \rExamAbortsDropped{} were objects dropped below the table plane; no exam episode was stopped for an invalid pose request or a joint limit. A 100-episode exam is a coarse instrument for events at this rate: zero aborts in 100 bounds the per-episode abort probability only to \rAbortUpperZero\% (Clopper--Pearson 95\%), one abort to \rAbortUpperOne\%, and a skill whose true abort rate is 1\% passes a zero-abort rule with probability \rPZeroOne\% (2\%: \rPZeroTwo\%; 3\%: \rPZeroThree\%). Two \texttt{place\_phone\_stand} campaigns illustrate the consequence: the first passed with 0/100 and the third failed with 1/100, and their abort-rate intervals overlap almost entirely.

To see whether the failures were sampling luck, the exam versions of the four campaigns nearest the bar were re-run on \rExtraEvalN{} further layouts (\cref{tab:extra}; report-only). All four scored 96--97\% success, but their abort rates separated into three tiers: essentially zero, about 1\%, and 3--4\%. The exam had scored them 1, 1, 4 and 2 aborts, in that order. The two \texttt{stack\_bowls\_two} verdicts were therefore correct, the \texttt{place\_phone\_stand} verdict was probably unlucky, and \texttt{hanging\_mug} sits where a binary zero-abort rule cannot decide. Repair did not remove these tail events: campaigns with large RSI deltas still aborted in the exam (\texttt{hanging\_mug} reached 42/42 development and 43/43 validation and aborted once in 100; \texttt{stack\_bowls\_two} 3 reached 47/47 and aborted four times), which is expected, since development and validation blocks of 36--47 layouts cannot reveal a 1--3\% event often enough for the author to act on it. For each of the \rnFrozen{} frozen skills, its 100 exam episodes and its caller-check episodes contain no abort, which bounds that skill's abort rate only to about \rAbortUpperOneTwenty\%; the skill that later aborted in reuse (\cref{sec:q3}) sits inside that bound. The second experiment tested the obvious remedy: with \rFEightDevNMin{} usable development layouts a 3\% abort rate appears about four times per version, yet the three campaigns still aborted \rFEightAborts{} times in their exams (\rFEightAbortWorkspace{} workspace exits, \rFEightAbortDropped{} drops). More layouts moved the exam scores to the top of the task's range but did not clear the gate. The aborts were not hidden from the author this time: the versions that went to the exam had aborted \rFEightDevAbortsExam{} times on the \rFEightDevNMin{} development layouts (exam: \rFEightAborts), every repair-round version aborted at least \rFEightDevAbortsMinRound{} times there, and the starting versions had aborted \rFEightDevAbortsStart{} times. The author read these aborts in its feedback each round, reduced them from \rFEightDevAbortsStartMax{} to 4 per 128 in one campaign, and never removed them.

\begin{table}[!tp]
\centering
\caption{Re-evaluation of the four campaigns nearest the bar on \rExtraEvalN{} new layouts (report-only). Success carries a Wilson interval, aborts a Clopper--Pearson interval.}
\label{tab:extra}
\footnotesize\setlength{\tabcolsep}{5pt}
\begin{tabular}{@{}lcccccc@{}}
\toprule
Campaign & Exam & Aborts & Success /300 & 95\% CI & Aborts & 95\% CI \\
\midrule
\tabExtraEvalRows
\bottomrule
\end{tabular}
\end{table}

\subsection{Q2a: in-session self-testing shows no detectable effect}
\label{sec:insession}

Every first draft in \cref{tab:compile} was written in a session in which the author could run practice episodes and its own simulator experiments. \Cref{tab:writeonce} compares the validation baselines of those drafts with drafts written under the two ablation conditions on the same scene (write once on all six tasks, practice only on four). Removing the simulator account (practice only) and removing practice altogether (write once) changed neither the first-draft development scores nor the validation baselines detectably: the task-stratified difference between the default condition and write-once is \rWOBaseHostOnceDiff{} points on the validation baseline (permutation $p = \rWOBaseHostOnceP$, one-sided for the default being better) and \rWODraftsHostOnceDiff{} points on development scores ($p = \rWODraftsHostOnceP$); practice-only versus write-once is \rWOBaseNoHostOnceDiff{} ($p = \rWOBaseNoHostOnceP$). Write-once drafts reached \rWOOnceBaseMin--\rWOOnceBaseMax\% on validation from the task card, the official environment source and the API alone. Very poor drafts occurred with and without the ability to run the package (\texttt{move\_can\_pot} 13/45 and \texttt{stack\_bowls\_three} 11/43 with full practice; \texttt{hanging\_mug} 6/42 write-once), and the two-draft selection absorbed them in each case. With one write-once campaign per task, this is evidence of no detectable effect rather than of none: the within-task spread of default baselines (SD \rWOWithinSD{} points over \rWOWithinDF{} degrees of freedom) gives the stratified difference a standard error of about \rWOSE{} points, so a one-sided test at 80\% power would detect only a difference of about \rWOMDE{} points or more. Within that limit, the in-session term of the decomposition in \cref{sec:terminology} is small for this author, and the measurable self-improvement in a campaign is the cross-session term.

\begin{table}[!tp]
\centering
\caption{Validation baseline of the selected first draft (\%, mean over campaigns; count in parentheses) by author condition. Default: practice episodes and own simulator account. Practice only: no account. Write once: the package cannot be run before delivery.}
\label{tab:writeonce}
\footnotesize\setlength{\tabcolsep}{6pt}
\begin{tabular}{@{}lccc@{}}
\toprule
Task & Default & Practice only & Write once \\
\midrule
\tabWriteonceRows
\bottomrule
\end{tabular}
\end{table}

\subsection{Q2b: repair beats rewriting on hidden layouts, and most of the gap is inheritance}
\label{sec:q2}

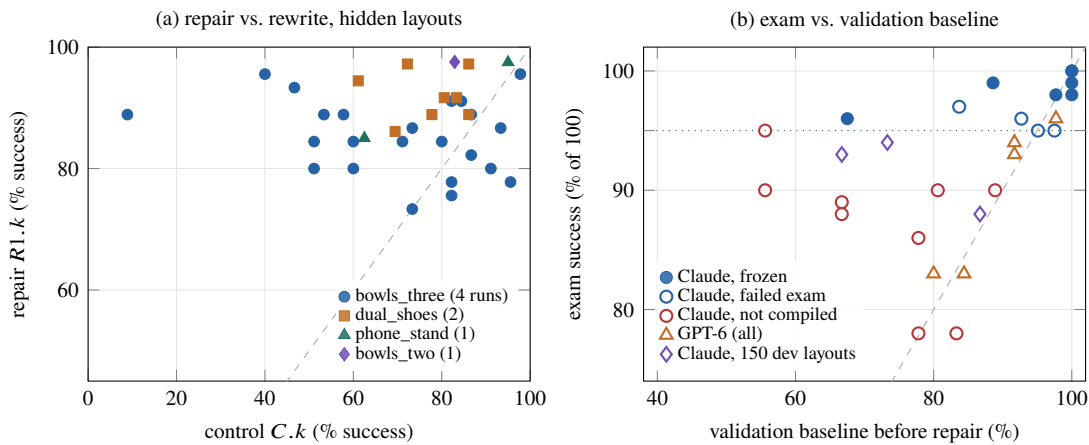
\begin{figure}[!tp]
\centering
\input{figures/learning.tex}
\caption{(a) Each point is one repair round of a controlled campaign: the repair version against the control draft of the same round, both on the hidden validation layouts; points above the diagonal favor repair. (b) Exam success against the hidden-validation baseline for all \rnCampaigns{} campaigns of both products and the three 150-layout campaigns; the dotted line is the 95\% bar.}
\label{fig:learning}
\end{figure}

\Cref{tab:heldout} and \cref{fig:learning}a give the per-round comparison on the \rHOCampaigns{} controlled campaigns that needed repair. Each repair version $R1.k$ and each control draft $C.k$, written in sessions of the same length limit from the same inputs except for history and evidence, was run once on the same hidden validation layouts (\rHOLayoutsMin--\rHOLayoutsMax{} per task; about \rHOEpisodesApprox{} episodes in all). Repair was better in \rHOPos{} of \rHORounds{} rounds, worse in \rHONeg{} and tied in \rHOTie, by \rHOMeanDiff{} layouts (\rHOMeanDiffPP{} points) per round on average; per campaign the mean difference was positive in \rHOCampPos{} of \rHOCampaigns{} (one-sided sign test $p = \rHOSignP$; exact sign-flip permutation on campaign means \rHOPermMean{} layouts, $p = \rHOPermP$). The same comparison on the development layouts, where the repair side has an advantage, gives \rDevAllPos{} of \rDevAllRounds{} rounds better (\rDevAllMeanDiff{} layouts), so the advantage is not an artifact of tuning to the development block.

Two further readings guard against the asymmetry noted in \cref{sec:control}. Relative to each campaign's own starting point on the same hidden layouts, the repair lineage averaged above the start in \rHORepairAboveStart{} of \rHOCampaigns{} campaigns (mean \rHORepairStartMean{} layouts; sign test $p = \rHORepairStartSignP$), while the control drafts averaged below it in \rHOControlBelowStart{} of \rHOCampaigns{} (mean \rHOControlStartMean), which is what a single draw against a selected best-of-two predicts; part of the per-round gap is therefore the starting point that repair keeps and a rewrite forfeits. The selection-matched best-of-$k$ comparison of \cref{eq:learn} is much closer: on the development layouts, $\mathrm{Learn}(k)$ at the end of R1 was \rLearnKList{} layouts (positive in \rLearnKPos{} campaigns, zero in \rLearnKZero, negative in \rLearnKNeg), and on the hidden layouts each side's best version favors repair in \rHOBestRepair{} campaigns and the control in \rHOBestControl. Independent rewrites have enormous variance, from \rHOControlMin{} to \rHOControlMax{} within one campaign, so a handful of them occasionally contain a version as good as the repair lineage's best. What repair buys is not a higher ceiling than lucky rewriting but reliability per round.

\begin{table}[!tp]
\centering
\caption{Per-round held-out comparison. Start is the selected R0 draft on the hidden validation layouts; each pair is repair $R1.k$ / control $C.k$ on the same layouts; the last column is each side's best version.}
\label{tab:heldout}
\footnotesize\setlength{\tabcolsep}{5pt}
\begin{tabular}{@{}llc>{\centering\arraybackslash}p{0.5cm}>{\raggedright\arraybackslash}p{4.2cm}cc@{}}
\toprule
Task & Run & Start & $k$ & $R1.k$ / $C.k$ & Mean $\Delta$ & Best \\
\midrule
\tabHeldoutRows
\bottomrule
\end{tabular}
\end{table}

\begin{figure}[!tp]
\centering
\input{figures/inherit.tex}
\caption{Second experiment: successes of each round's repair, inherit-only and rewrite versions on the \rFEightHOLayouts{} hidden validation layouts; the dashed line is the starting version. Inherit-only and repair rarely fall below the start; rewrites vary from \rFEightControlMin{} to \rFEightControlMax.}
\label{fig:inherit}
\end{figure}
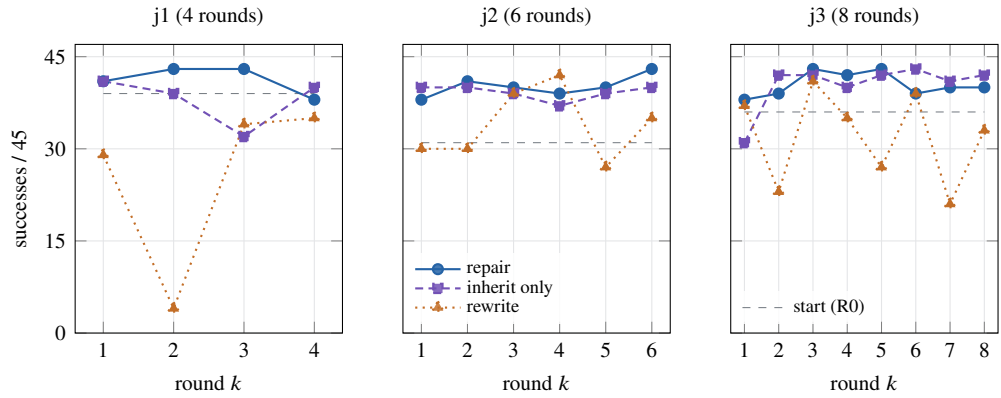

The second experiment separates the two things a repair session has that a rewrite does not (\cref{tab:feightho}, \cref{fig:inherit}). On the \rFEightHOLayouts{} hidden layouts, repair exceeded the from-scratch rewrite by \rFEightRCMean{} layouts per round on average (per campaign \rFEightRCPerCamp); the inherit-only lineage, which revises the previous version with its own practice runs but without the runtime's development-layout feedback, exceeded the rewrite by \rFEightICMean{} (\rFEightICPerCamp); and repair exceeded inherit-only by \rFEightRIMean{} (\rFEightRIPerCamp). All three differences had the same sign in all \rFEightN{} campaigns, which with three campaigns as the unit is the smallest attainable one-sided sign-test $p$ of \rFEightRCSignP; the per-round counts (\rFEightRCPos/\rFEightRCNeg/\rFEightRCTie, \rFEightICPos/\rFEightICNeg/\rFEightICTie{} and \rFEightRIPos/\rFEightRINeg/\rFEightRITie{} better/worse/tied for the three differences) are descriptive, since rounds within a lineage are not independent, and none of these results is claimed as significant. Within those limits the answer to Q2 is: on layouts none of the lineages has seen, revising the previous version beats rewriting from scratch, most of the gap comes from inheriting the previous version across sessions, and the runtime's development-layout feedback adds a little more at the margin. Both are components of self-improvement in the sense of \cref{sec:terminology}; neither is selection, which all three lineages share. The mechanism is consistent with this reading: of the \rMechSessions{} repair sessions of the primary author, \rMechReadFeedback{} read the per-layout feedback of the previous version, \rMechReadVerdict{} read the comparison verdict, \rMechReadNotes{} read the author's own notes, and \rMechCites{} (\rMechCitesPct\%) cite specific failure classes, scores or layout seeds in the delivered rationale; the development feedback is read and used, and its marginal value over inheritance and the author's own practice on this task is small. The remaining gap between repair and the gate is the one described in \cref{sec:safety}: on the hard tasks, repair drove development scores to within a few layouts of the target and validation to about 90\%, and the exam then found aborts that the development block showed too rarely to act on.

\begin{table}[!tp]
\centering
\caption{Second experiment, per-round held-out comparison on the \rFEightHOLayouts{} hidden validation layouts: successes of the repair ($R1.k$), inherit-only ($I.k$) and rewrite ($C.k$) versions of each round, and the campaign-mean differences repair$-$control (R$-$C), inherit$-$control (I$-$C) and repair$-$inherit (R$-$I). With three campaigns the one-sided sign test cannot go below $p = 0.125$ (\cref{tab:feightcmp}).}
\label{tab:feightho}
\footnotesize\setlength{\tabcolsep}{5pt}
\begin{tabular}{@{}lc>{\raggedright\arraybackslash}p{2.6cm}>{\raggedright\arraybackslash}p{2.6cm}>{\raggedright\arraybackslash}p{2.6cm}ccc@{}}
\toprule
Run & Start & $R1.k$ & $I.k$ & $C.k$ & R$-$C & I$-$C & R$-$I \\
\midrule
\tabFEightHeldoutRows
\bottomrule
\end{tabular}
\end{table}

\begin{table}[!tp]
\centering
\caption{Second experiment, the three pre-registered differences: per-campaign means (layouts of \rFEightHOLayouts), their mean, per-round counts (better/worse/tied; descriptive only) and the campaign-level sign test. ``Feedback'' is the runtime's development-layout evaluation; all three lineages keep their practice runs.}
\label{tab:feightcmp}
\footnotesize\setlength{\tabcolsep}{5pt}
\begin{tabular}{@{}>{\raggedright\arraybackslash}p{2.3cm}>{\raggedright\arraybackslash}p{2.9cm}>{\centering\arraybackslash}p{2.0cm}ccc>{\centering\arraybackslash}p{0.9cm}@{}}
\toprule
Difference & Measures & Per campaign & Mean & Rounds & Camp.\ & $p$ \\
\midrule
\tabFEightCompareRows
\bottomrule
\end{tabular}
\end{table}

\subsection{A second author product}
\label{sec:gpt}

The five campaigns with \rGPTProduct{} (\cref{tab:compile}, lower block) reproduce the pattern without a control group. First drafts started at \rGPTBaseMin--\rGPTBaseMax\% on validation, \rGPTRepairsMin--\rGPTRepairsMax{} repair rounds moved development scores to 38--43 of 42--44, and the exams scored \rGPTExamMin--\rGPTExamMax{} with \rGPTAbortsList{} aborts (in the order of \cref{tab:compile}); \rGPTRateBar{} of five reached the success bar and none passed the gate: four aborted at least once and the fifth scored 93, for the same two reasons as the primary author on the same tasks (success of 83 on the three-object stack and 93--94 on the shoe task; a single abort on \texttt{hanging\_mug} at 96). RSI deltas were small (\rGPTDeltaMin{} to \rGPTDeltaMax) because the starting points were high. The product's subscription quota ran out twice; the runtime paused and resumed the session's remaining time without counting a round. All five audits were clean. These campaigns show that the protocol transfers to another agent product; they are not a comparison of the two products' strength, which would need matched tasks and budgets.

\subsection{Q3: reuse by executors that did not author the skill}
\label{sec:q3}

\Cref{tab:reuse} and \cref{fig:reuse}a report \protoB on three frozen skills (\texttt{move\_can\_pot} 3, \texttt{place\_phone\_stand} 1 and \texttt{stack\_bowls\_two} 1 of \cref{tab:compile}), each on 50 layouts disjoint from every compile-time seed. With no model at all the skills succeeded in \rPoolF{} episodes (\rPoolFPct\%; Wilson \rPoolFCI\%); with the DeepSeek-flash caller reading only \texttt{SKILL.md}, in \rPoolO{} (\rPoolOPct\%; \rPoolOCI\%). The two arms differ by one episode in 150 (on the shared seeds, \rBDiscFonly{} layout succeeded only with code, \rBDiscOonly{} only with the caller, and \rBBothFail{} failed under both), each skill's reuse score matches its exam score, and the only safety aborts occurred on \texttt{stack\_bowls\_two} (\rReuseStackBowlsTwoFAborts{} with code, \rReuseStackBowlsTwoOAborts{} with the caller), the task whose two other campaigns the 300-layout re-evaluation in \cref{sec:safety} placed at 3--4\%. The caller made \rOdsDecisions{} decisions over the 150 episodes, none invalid: \rReuseMoveCanPotODec{} per episode on \texttt{move\_can\_pot} and \texttt{stack\_bowls\_two}, where the standard flow in \texttt{SKILL.md} is a single plan, and \rReusePlacePhoneStandODec{} on \texttt{place\_phone\_stand}, where a failed step in the first plan is followed by the recovery branch documented in the description. Each decision read about \rOdsInTokensMean{} input tokens and produced about \rOdsOutTokensMean, with median latency \rOdsLatencyMedian{} s (90th percentile \rOdsLatencyPNinety{} s). The capability therefore resides in the package: an executor with no model matches the author's exam, and a model from another family, given only the description, follows it correctly.

\begin{table}[!tp]
\centering
\caption{\protoB on three frozen skills, 50 new layouts per skill, both arms on the same layouts. Compile is the author cost from the manifest; exam is the skill's own qualification score; cost is the caller's model cost per episode from provider usage and the captured price; amortized columns are $C/N + c_X$ at $N{=}100$ and $N{=}1000$ deployments.}
\label{tab:reuse}
\scriptsize\setlength{\tabcolsep}{4pt}
\begin{tabular}{@{}lcclccccccc@{}}
\toprule
Skill & Compile \$ & Exam & Arm & Success & Wilson 95\% (\%) & Aborts & Decisions & Model cost / episode \$ & At 100 \$ & At 1000 \$ \\
\midrule
\tabReuseRows
\bottomrule
\end{tabular}
\end{table}

\subsection{Q4: cost per episode}
\label{sec:q4}

\begin{figure}[!tp]
\centering
\input{figures/reuse.tex}
\caption{(a) Success of each frozen skill in its exam and under the two \protoB executors. (b) Per-episode cost with compilation amortized over $N$ deployments; the dotted lines at \$\rSideDstCost{} and \$\rSideDdsCost{} are the strong model and DeepSeek controlling the robot without a skill in the runtime self-test (two episodes per arm), an order-of-magnitude reference and not a measured baseline.}
\label{fig:reuse}
\end{figure}
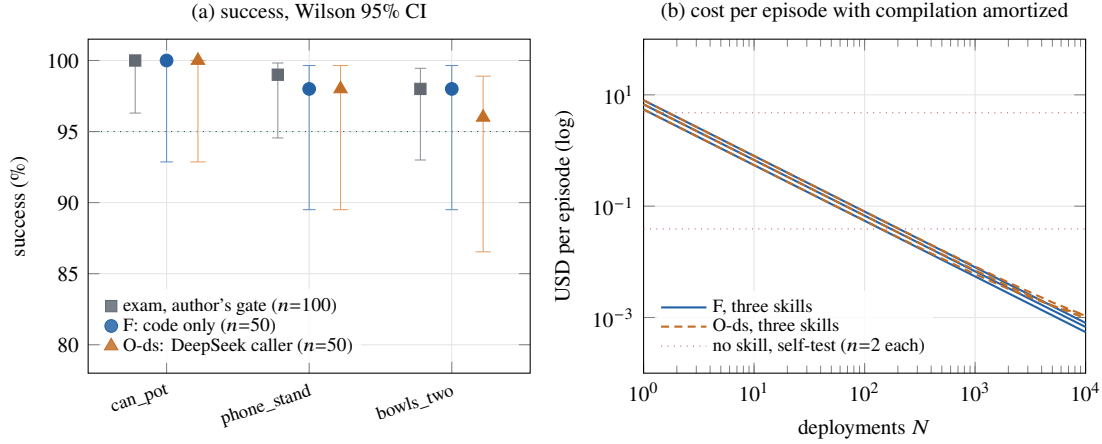

Compiling the three skills cost \$\rReuseCompileTotal{} of author budget in total (\$\rReuseCompileMin--\rReuseCompileMax{} each). That is the cost of the campaigns that succeeded; the program that produced the seven frozen skills cost \$\rAllUSDTotal{} over \rnClaude{} campaigns, or \$\rProgramUSDPerFrozen{} per frozen skill and \$\rProgramUSDPerTask{} per compiled task, and the second experiment added \$\rFEightCostTotal{} for three campaigns that did not compile. Simulation is not priced: the \rYieldCampaigns{} compile campaigns occupied \rYieldSlotHours{} simulator-slot hours on one GPU (\cref{sec:other}), which at a rental price of one to two dollars per GPU-hour would add an amount comparable to the author bill; the runtime's own engineering is a fixed cost we do not amortize. Executing a frozen skill has no model cost per episode with code and \$\rOdsCostMin--\rOdsCostMax{} with the caller; the 150 caller episodes cost \$\rOdsTotalCost{} together. With compilation amortized (\cref{fig:reuse}b), the per-episode cost of the most expensive skill with the caller is \$\rAmortHundredMax{} at 100 deployments and \$\rAmortThousandMax{} at 1000, below one cent. We do not report a break-even against direct online control, because the direct-control arms of the full design were not run as a formal evaluation. The only evidence is a runtime self-test of two episodes per arm on \texttt{move\_can\_pot} with the same prompt template and a package that exposes only the primitives: the primary author's model, controlling the robot directly, succeeded in \rSideDstSuccess{} of \rSideDstN{} episodes at \$\rSideDstCost{} per episode and \rSideDstDec{} decisions (about \rSideDstOutTokens{}k output tokens per episode), and DeepSeek-flash in \rSideDdsSuccess{} of \rSideDdsN{} at \$\rSideDdsCost. These four episodes fix an order of magnitude, not a rate: per-episode reasoning from primitives costs dollars with the strong model and cents with the cheap one, and in this sample completed nothing; a compiled skill costs a fraction of a cent and completes 98\%. The order of magnitude agrees with the one public figure we know of for per-episode reasoning with evolved experience: EmbodiedEvo reports a mean model-API cost of \$5.73 per test episode for a GPT-6 Astra agent carrying its Skills and Lessons, excluding the playground learning phase~\cite{li2026embodiedevo}. That figure comes from different tasks, a different simulator and a different observation interface, is not a measurement of ours, and does not enter any break-even computation; it only says that the per-episode cost of the route we compare against is dollars, not cents.

\subsection{Caller check, audit, and the episode ledger}
\label{sec:other}

The post-freeze caller check with the author product scored \rCCSuccess{} of \rCCValid{} valid episodes over the seven frozen skills (\rCCRate\%; two of the \rCCScheduled{} scheduled episodes failed on infrastructure in every attempt), with \rCCDecisionsMin--\rCCDecisionsMax{} decisions per episode; no skill approached the warning threshold. The isolation audit graded \rIsoClean{} of \rnClaude{} primary-author campaigns clean, \rIsoMetadata{} metadata and \rIsoReview{} review; the three non-clean grades record that another campaign's identifier or home path appeared in a process or directory listing, with no file content read. Process listings are visible across users on both machines; this is a known gap of the isolation, not a contamination, and the three campaigns are retained. The second author's five audits were clean. In the second experiment two campaigns were clean and one (j2) was graded review for the same reason, another campaign's directory name and identifiers in a process listing with no content read; it was reviewed and retained, and removing it does not change the sign of any of the three differences in \cref{tab:feightcmp}.

Every episode of both protocols is in the hash-chained ledger with its seed, version, mode, outcome, failure class, safety events, primitive calls, simulated and wall-clock time, and the entry and exit marks of each atom call, so that the primitive trajectory is segmented by sub-goal and labelled with the atom's name and its \texttt{ok} value. The \rYieldCampaigns{} compile campaigns alone recorded \rYieldEpisodesAll{} episodes (\rYieldEpisodesProtocol{} in the protocol itself, of which \rYieldEpisodesPractice{} were the author's practice runs, and \rYieldEpisodesPostHoc{} in the report-only re-runs); \rYieldEpisodesInfra{} (\rYieldInfraPct\%) were infrastructure failures and are excluded, leaving \rYieldEpisodesValid{} labelled trajectories, \rYieldSuccess{} successful and \rYieldFail{} failed with a cause (\rYieldFailFinished{} finished without satisfying the predicate, \rYieldFailWorkspace{} workspace exits, \rYieldFailDropped{} drops, \rYieldFailAtomError{} atom errors, \rYieldFailInvalidQuat{} invalid pose requests, \rYieldFailBudget{} budget exhaustion). Every fallback episode is segmented: a trajectory contains \rYieldSegmentsMean{} atom calls and \rYieldSubgoalsSucc{} top-level sub-goals on average (\rYieldSubgoalsFail{} in failed episodes), \rYieldCoverageFull\% of fallback episodes have every primitive call inside an atom, and none bypasses the atoms. A valid trajectory took a median \rYieldWallMedianSec{} s of simulator-slot time, so the study produced \rYieldPerSlotHour{} labelled trajectories per slot-hour (\rYieldSlotHours{} slot-hours in all) and \rYieldPerWallHour{} per campaign wall-clock hour with four slots. These traces carry sub-goal boundaries, language (from \texttt{SKILL.md}), success labels and failure types, which teleoperation data lack, but not control-frequency sensor streams; their value as training data for embodied models is not tested here (\cref{sec:discussion}).

\begin{table}[!tp]
\centering
\caption{Published results on the same seven tasks, for reference only. The twelve columns under ``published'' are transcribed from the RoboTwin~2.0 paper~\cite{chen2025robotwin2} (arXiv:2506.18088v2): five policies trained from 50 clean demonstrations per task with single-task fine-tuning (RDT and $\pi_0$ from pretrained weights), evaluated over 100 rollouts in the Easy setting (clean scene) and the Hard setting (cluttered table with randomized lighting, textures and table height), from its Appendix~K, Table~10; its LLM-generated task programs (DeepSeek-V3 with multimodal feedback, ``R2.0 + MM FB''), which read privileged object poses at run time and whose metric is the mean success rate over candidate programs rather than one 100-episode run, from its Appendix~G.3, Table~9; and the official expert program on the Aloha embodiment, from its Appendix~L, Table~11 (RoboTwin~2.0 column). Table numbers in the header refer to that paper. None of these were measured in this paper and all were obtained under other conditions: the frozen skills use no demonstrations and no privileged poses, perceive from RGB-D with the camera calibration, run in the clutter tier (clean configuration plus up to ten distractors, fixed lighting, textures and table height; between Easy and Hard), and must also pass the zero-safety-abort gate and the 2\,s settle before the predicate. The table is therefore context, not a comparison, and no differences are computed. ``Expert, ours'' is the official expert program run ten times in phase 0 on the task's fixed phase-0 layouts (the same ten layouts for every campaign of a task), given as the range over the \rFNinePZeroCampaigns{} clutter-tier campaigns in the export (both author products and the second experiment; campaign count in parentheses), not pooled. ``Exam'' repeats \cref{tab:compile,tab:feight}: Claude Code formal campaigns in table order with the number frozen over the number of campaigns, then the GPT-6 campaigns and the three 150-layout campaigns of the second experiment.}
\label{tab:published}
\scriptsize\setlength{\tabcolsep}{3.0pt}
\begin{tabular}{@{}l ccccc ccccc c c c >{\raggedright\arraybackslash}p{3.4cm}@{}}
\toprule
& \multicolumn{12}{c}{Published in RoboTwin~2.0 (not measured in this paper; success \%)} & \multicolumn{2}{c}{This paper, clutter tier (measured)} \\
\cmidrule(lr){2-13}\cmidrule(lr){14-15}
& \multicolumn{10}{c}{Demonstration-trained policies (their Table~10)} & LLM code & Expert & Expert, ours & Exam, of 100: Claude Code formal (frozen/campaigns); \\
\cmidrule(lr){2-11}
& \multicolumn{5}{c}{Easy} & \multicolumn{5}{c}{Hard} & (their Table~9) & (their Table~11) & (of 10) & GPT-6; 150 development layouts \\
\cmidrule(lr){2-6}\cmidrule(lr){7-11}
Task & RDT & $\pi_0$ & ACT & DP & DP3 & RDT & $\pi_0$ & ACT & DP & DP3 & & & & \\
\midrule
\tabFNineRows
\bottomrule
\end{tabular}
\end{table}

\subsection{Published results on the same tasks, for reference}
\label{sec:published}

\Cref{tab:published} places the exam results beside the numbers that the RoboTwin~2.0 paper reports on the same seven tasks, simulator and embodiment~\cite{chen2025robotwin2}: five policies trained from 50 clean demonstrations per task (its Table~10, Easy and Hard settings; version v2 of arXiv:2506.18088), its own LLM-generated task programs (its Table~9), and the official expert program (its Table~11). All five policies and all seven tasks are listed. None of these numbers were measured here, and the conditions differ in data, run-time information, scene and metric, as the caption lists; the table is context, not a comparison, and we compute no differences and draw no ranking from it. Three directional observations are all we take from it. First, the seven tasks are not easy for demonstration-trained policies: \rFNineEasyBelowSixty{} of the \rFNineEntries{} Easy entries are below 60\% and \rFNineHardBelowTwentyFive{} of the \rFNineEntries{} Hard entries are below 25\%. Second, the LLM-generated code of RoboTwin~2.0, which reads privileged object poses, scores 0 on \rFNineCodegenZero. Third, the skills compiled here, with no demonstrations and no privileged poses, scored \rFNineExamMin--\rFNineExamMax{} of 100 on exams drawn after freezing in a scene tier that lies between Easy and Hard, at \$\rFrozenUSDMin--\rFrozenUSDMax{} of author spend per frozen skill (\$\rProgramUSDPerFrozen{} at program level, \cref{sec:q4}), where the demonstration route spends 50 demonstrations and a fine-tuning run per task; the two cost accountings are not convertible into each other. The one entry measured under our own conditions is the official expert program in phase 0: on \texttt{stack\_bowls\_three} it succeeded in \rFNinePZeroStackBowlsThree{} of 10 episodes across \rFNinePZeroNStackBowlsThree{} campaigns, while the compiled skills for that task scored \rSBThreeAllExamMin--\rSBThreeAllExamMax{} on the exam and were stopped by the safety rule, not by task completion (\cref{sec:safety}).

\section{Discussion}
\label{sec:discussion}

\paragraph{Skills before scale.}
The results test the precondition of the hypothesis stated in the introduction, not the hypothesis. A strong general model compiled four of seven tasks into skills that passed a qualification exam on layouts drawn after freezing, for under ten dollars each; the skills ran on new layouts without that model at the same success, and a model from another family could drive them from the description alone for a fraction of a cent per episode. What the experiments do not show is that skill execution data improve an embodied model, nor that any of this transfers to physical hardware. If both hold, the order of the language-model history can be inverted for robots: the strong general model exists now, skills compiled from it can put robots to work in bounded scenes now, and the labelled trajectories that execution produces are the corpus that embodied foundation models lack. If either fails, compiled skills remain a deployment technique with a measured cost profile.

\paragraph{What recursion bought, and what it did not.}
Five of the seven frozen skills needed no repair: on the easy tasks the first draft was already at the bar, and the protocol's contribution was qualification and freezing, not improvement. Recursion mattered on the other tasks, and there its effect is characterized by the controls: on layouts none of the lineages has seen, a session that revises the previous version delivers a better version than a session that starts over, in \rHOCampPos{} of \rHOCampaigns{} and \rFEightN{} of \rFEightN{} campaigns, and the inherit-only lineage shows that most of that advantage is carried by the artifact itself, by inheriting the previous version across sessions, with the runtime's development feedback adding a smaller increment on the task where we measured it. Both are self-improvement of the skill in the sense of \cref{sec:terminology}; the finding says where in the loop the improvement lives. What recursion did not do is carry a hard task over the gate: \texttt{stack\_bowls\_three} failed on safety aborts in all \rSBThreeAllCampaigns{} campaigns, with 43 and with \rFEightDevNMin{} usable development layouts, even as the RSI delta reached \rDeltaMaxRepaired{} points. The larger block let the author see the aborts (\rFEightDevAbortsExam{} on 128 development layouts for the versions it sent to the exam, at least \rFEightDevAbortsMinRound{} for every repair-round version) and did not let it remove them, so the hypothesis that the loop merely lacked evidence is not supported by this experiment; the bottleneck is now more plausibly the strategies the author finds for this task and the kernel they run on than the amount of feedback. One hypothesis, not a result, follows from the kernel's design: the planner does not see the objects on the table (\cref{sec:setup}), and the dominant abort is an object pushed out of the workspace (\rExamAbortsWorkspace{} of \rExamAbortsTotal{} exam aborts across the 25 campaigns of both products, \rFEightAbortWorkspace{} of 16 in the second experiment), which is what an arm sweeping through a third bowl or a distractor would produce; a planner that treats detected objects as obstacles would test it. Abort-focused evidence between rounds, which we proposed after the first results, remains untested and is no longer the obvious next step.

\paragraph{Gates for rare events.}
A zero-abort rule over 100 episodes is a coin flip for skills whose true abort rate is near 1\%, and it passes a 3\% skill one time in twenty. The 300-layout re-evaluations separated tiers that the exam could not. A better gate would state the safety requirement as an upper confidence bound on the abort rate, or as a sequential test that stops early on clearly safe or clearly unsafe skills~\cite{wald1945sequential}; both cost more simulation than 100 episodes and both cost nothing in model calls, since the exam runs in fallback mode. Reporting the interval beside the verdict, as \cref{tab:compile} does, is the minimum. As a post-hoc observation, not a change to the pre-registered rule: under a success-only rule \rnRateBar{} of \rnClaude{} primary campaigns and \rGPTRateBar{} of five GPT-6 campaigns would have passed, and under a rule that bounds the abort rate by its upper confidence limit no campaign with an observed abort would, so the gate's verdicts are sensitive to how the safety requirement is written.

\paragraph{What is externalized.}
The executor crossings answer a question that success rates alone cannot: where the capability lives. If only the author's model could execute the skill, the package would be a reasoning scaffold and the capability would still be in the runtime. Here an executor with no model matches the author's exam, and a cheap caller from another family adds nothing on these tasks and loses nothing, because the descriptions rarely need interpretation: one plan per episode, or one recovery branch. The kernel keeps motion generation, collision checking and limits; the skill holds decomposition, sub-goal ordering, preconditions, postconditions, observation timing, bounded recoveries and permissions; the caller, when present, holds branch selection. Tasks whose recovery sections are long or ambiguous would load the caller more, and the gap between callers would then measure the description; we did not observe such a task among the three measured. EmbodiedEvo's finding that experience evolved with a strong model raises the success of two weaker models that read it~\cite{li2026embodiedevo} points in the same direction as our cross-family caller, with the difference that the weaker models there still reason in every episode, whereas here the description is enough for a caller to execute a package it never saw the code of.

\paragraph{Strategy substitution versus perception completion.}
Text corpora do not contain grip forces, and the author has no tactile input, yet skills for fragile or slippery objects can be compiled if the author discovers a strategy whose success does not depend on the missing quantity: supporting an object from below instead of pinching it, or re-grasping after a slip check, rather than estimating a safe force. Our tasks contain no such hidden property, so this is a hypothesis grounded in the mechanism rather than a result; RoboTwin~2.0 has no fragile or low-friction assets, and adding them would change the official code tree that the tamper check pins. The boundary is also clear: tasks that require identifying the property, such as sorting real from artificial eggs, cannot be solved by substitution, and there the modality-completion route~\cite{ye2026tacgen,higuera2024sparsh} is the only one.

\paragraph{When is an author worth paying for.}
The author is expensive per session and the executor is nearly free per episode, so the economics depend on reuse and on failure. The seven frozen skills cost under \$10 each; the nine primary campaigns that did not compile cost \$\rNCUSDMin--\rNCUSDMax{} each and the three of the second experiment \$\rFJOneUSD--\rFJThreeUSD, and all returned a ledger and a near-miss package, not a skill, so the cost of a program of compilation is dominated by the tasks that fail: \$\rProgramUSDPerFrozen{} per frozen skill in this program against \$\rFrozenUSDMedian{} for the median successful campaign. A task that a strong model solves at 95\% on the first draft needs only qualification and freezing; a task that it solves at 60\% needs repair and may still fail the safety rule. The break-even question against per-episode reasoning remains open for lack of a measured direct-control baseline; the self-test suggests the answer is a small number of deployments.

\paragraph{Threats to validity.}
The author may have seen RoboTwin~2.0 during pretraining; exam and reuse layouts are drawn after freezing, so memorization of public content cannot supply exam answers, but it can make first drafts stronger than they would be on a private task suite, and the write-once result should be read with that in mind. The task card includes the success-predicate source, so the author writes against the criterion; we consider that the realistic setting for a scene an integrator specifies, but it is a lenient one. The held-out comparison and the 300-layout re-evaluations were added after the first results were seen and are reported as such. Replication is unbalanced and small: one to six campaigns per task, eight controlled campaigns, three skills in \protoB with one cheap caller; intervals carry that uncertainty, and cross-task summaries are medians, not pooled rates. DeepSeek prices vary by time of day and the price snapshot is dated. Process listings were visible across isolated users. The author products are closed subscription services: their served models, system prompts and tools can change behind a fixed version string and cannot be archived, so \protoA outcomes are a dated measurement of two products rather than of a method that a reader can re-run, and only the \protoB episodes are replayable from stored responses. The second experiment deviated from its registration in execution (voided rounds after a provider login failure, a redrawn exam, operator restarts after connection drops; \cref{app:prereg}); none of the deviations changed a rule or a result after it was seen. Finally, the runtime version advanced during the study; the increments are listed with their dates and none touched the author prompt or a gate rule.

\section{Limitations and Safety}
\label{sec:limits}
The study is simulation-only; friction, compliance, calibration drift and human proximity produce failures a clean simulator does not, and no real-robot run is reported, whereas EmbodiedRSI reports real-robot results~\cite{song2026embodiedrsi}. Seven tasks on one scene tier is a small suite next to the 42 RoboDojo tasks of concurrent work~\cite{li2026embodiedevo}. Our author receives RGB-D images and camera calibration and writes its own perception, while EmbodiedEvo's agent works from RGB alone, so the two settings are not directly comparable in perception difficulty. Restricted Python trades expressiveness for verifiability and cannot express contact-rich dexterity. The primitive API has single-arm actions, so bimanual coordination is serialized. Passing an exam is empirical evidence about declared layouts, harness versions and safety envelopes, not a proof of physical safety; the exam's resolution for rare events is stated in \cref{sec:safety}. Tamper evidence detects changes; it does not prevent deliberate forgery, and session transcripts are retained for audit. The author reads the success predicate. \protoB ran its minimal form: no second or third caller family, no strong-model caller, no format smoke test, and no formal direct-control baseline, so non-inferiority margins and break-even counts from the full design are not reported. The inherit-only comparison rests on three campaigns of one task with one author product and is existence evidence; the inherit-only author still receives one bit of information per round (whether the head was replaced), keeps its own practice runs, and works from a fresh host account, and the two changes of the second experiment were introduced together. Process listings are visible across the isolated users on both machines, so another campaign's identifiers can be seen, though not its content. Meta-level improvement of the author's procedure is not claimed. Physical deployment requires isolation during testing, emergency stops, force and speed limits, independent monitoring and staged release, none of which the protocols supply.

\section{Conclusion}
We separated the authoring of robot skills from their execution and measured the separation. A strong general model, used only offline, produced \rnFrozen{} immutable skill packages on \rnTasksFrozen{} of \rnTasks{} RoboTwin~2.0 tasks that passed a qualification exam on layouts drawn after freezing, most of them from a first draft that needed only qualification, at \$\rFrozenUSDMin--\rFrozenUSDMax{} each and \$\rProgramUSDPerFrozen{} per skill when the failed campaigns are counted. The three tasks that did not compile were stopped by rare safety aborts that a 100-episode exam can barely resolve and that a larger development block did not remove. On hidden layouts, revising the previous version beat rewriting from scratch in every controlled campaign, and an inherit-only lineage located most of that advantage in the inherited artifact rather than in the feedback read about it; self-testing within a session showed no detectable effect on first drafts. A second author product behaved the same way. The frozen skills ran at \rPoolFPct\% on new layouts with no model and at \rPoolOPct\% with a caller from another family at a fraction of a cent of model cost per episode. The protocols, ledgers and seeds are released so that the same measurements can be repeated with other authors, executors and tasks. Whether the data these skills produce while working can, in turn, train the embodied models that do not yet exist is the question the next study should ask.

\section*{Data and Code Availability}
The compile and reuse runtimes, adapters, task configurations, prompts, seed generators, environment fingerprints, hash-chained ledgers, raw caller responses, analysis scripts and the frozen skill packages are released at \url{https://github.com/Orillusion/embodied-skill-rsi} (release v1.0): the code under Apache-2.0, the data, ledgers and frozen skills under CC BY 4.0. The released runtime is the 5.9.0 release build; it differs from the frozen build used in the experiments only in site configuration (\texttt{docs/runtime\_release\_vs\_frozen\_5.9.0.diff}). Author session transcripts, raw model inputs and outputs, and per-episode images are not released and are retained for audit. \texttt{python3 analysis/reproduce.py} rebuilds the main tables from the public ledgers and checks them automatically; it needs no GPU. Third-party assets and model weights remain under their original licenses.

\section*{Ethics Statement}
The experiments use simulated robots and commercial model products accessed through their ordinary subscription or API channels; no human participants, personal data or animals are involved. Any physical deployment of compiled skills requires institutional safety review.

\section*{Generative AI Use}
Generative AI systems were used to support literature organization, language drafting, figure scaffolding and protocol structuring, and, as the object of study, to author skills. The runtime, the analysis scripts and the released repository were written and tested with the assistance of a generative AI coding product (Claude Code) and were reviewed and verified by the authors. All technical claims, citations, methods, analyses, code and final text were verified and approved by the human authors, who retain full responsibility.

\setlength{\bibsep}{3pt plus 2pt minus 2.5pt}
\bibliographystyle{unsrtnat}
\bibliography{references}

\clearpage
\appendix
\raggedbottom

\section{\fs Package and Primitive API}
\label{app:package}

\begin{table}[H]
\centering
\caption{Primitive API exposed by the kernel. ``Cost'' is the charge against the 90-call episode budget. Every \texttt{move\_pose} and \texttt{gripper} call is an independent planning request; the official plan-success latch is reset before each call.}
\label{tab:api}
\small
\begin{tabular}{@{}llll@{}}
\toprule
Primitive & Signature & Returns & Cost \\
\midrule
\texttt{task\_spec} & \texttt{()} & instruction, roles, criterion source & 0 \\
\texttt{observe} & \texttt{()} & ok, one component per role, table-height estimate & 1 \\
\texttt{ee\_pose} & \texttt{(arm)} & $[x, y, z, q_w, q_x, q_y, q_z]$ & 0 \\
\texttt{gripper\_state} & \texttt{(arm)} & commanded, measured, open & 0 \\
\texttt{move\_pose} & \texttt{(arm, pose7)} & bool (planned and executed) & 1 \\
\texttt{gripper} & \texttt{(arm, value in [0,1])} & bool & 1 \\
\texttt{wait} & \texttt{(seconds $\le 2$)} & bool & 1 \\
\bottomrule
\end{tabular}
\end{table}

\noindent The manifest written at freezing has the following fields (values from the \texttt{move\_can\_pot} skill of campaign 3, abbreviated):
\begin{lstlisting}[style=skill]
{"protocol": "skill-rsi-compile-v5", "runtime_version": "5.5.2", "api_contract": "primitives-v4",
 "scene": {"name": "clutter", "task_config": "demo_clean", "overrides": {"cluttered_table": true, ...}},
 "environment": {"gpu": "NVIDIA GeForce RTX 5090", "driver": "595.71.05", "torch": "2.8.0+cu128", ...},
 "env_fingerprint": "<sha256>",
 "gate": {"mode": "fallback", "success_rate_min": 0.95, "exam_n": 100, "exam_success": 100,
          "exam_valid": 100, "outcome": "passed", "decided_at": null, "episodes_run": 100,
          "speed": {"passed": true, "median": 7.83, "limit": 16.1}, "regression": {"passed": true, "worse_than": []}},
 "baseline_val": {"candidate": "R0.a", "success": 44, "valid": 44, "rate": 1.0},
 "caller_check": {"n": 20, "success": 20, "valid": 20, "decisions_mean": 1.0, "rejected_decisions": 0, "warning": null},
 "dev_seeds": [70000, ...], "val_seeds": [71000, ...], "exam_seeds": [...100...], "caller_check_seeds": [...20...],
 "compile_cost": {"rounds": 0, "author_sessions": 2, "author_usd": 7.9905, "author_hours": 1.752,
                  "sim_episodes": {"tryrun": 56, "infra_failed": 1, "dev": 90, "val": 44, "exam": 100},
                  "wallclock_hours": 6.18},
 "author": {"adapter": "claude_code", "product": "Claude Code", "version": "2.1.284 (Claude Code)", "model": "opus", "effort": "high", ...},
 "files": {"SKILL.md": "<sha256>", "skills.py": "<sha256>", "perception.py": "<sha256>"}}
\end{lstlisting}

\noindent Restricted-Python rules for \texttt{skills.py} and \texttt{perception.py}: no \texttt{import}; no \texttt{while}, \texttt{try}, \texttt{class} or \texttt{lambda}; a builtin allowlist (arithmetic, \texttt{len}, \texttt{range}, \texttt{min}, \texttt{max}, \texttt{abs}, \texttt{sorted}, \texttt{enumerate}, \texttt{zip}, list and dict literals); no names beginning with an underscore; a static source-line limit. The validator is the same file for every author and every executor.

\section{Gate Parameters and Qualification Checklist}
\label{app:gate}

\begin{table}[H]
\centering
\caption{Parameters of \protoA and \protoB as run. All \protoA values were fixed before the first campaign; the two report-only evaluations were added during the study at the versions shown.}
\label{tab:params}
\small
\begin{tabular}{@{}p{3.6cm}p{3.6cm}p{8.4cm}@{}}
\toprule
Parameter & Value & Rationale \\
\midrule
Scene tier & clutter (\texttt{demo\_clean} + up to 10 distractors; fixed lighting; no table-height randomization) & the fully randomized tier was too easy in a pilot (first drafts above 82\%); table-height randomization conflicts with absolute-height success predicates \\
R0 drafts & 2 & independent starting points differ more than one lineage's repairs \\
Development layouts & 50 seeds (70000--70049); \rDevNMin--\rDevNMax{} usable & separates 85\% from 95\% versions; layouts the generator cannot stabilize are dropped for every version \\
Validation layouts & 50 seeds (71000--71049); \rValNMin--\rValNMax{} usable & hidden; used for the baseline, the target check and the held-out re-run; never fed back \\
Development / validation target & $\geq 96\%$ / $\geq 94\%$ of usable & first integers at or above 95\% on 50, one layout of slack against a hidden set \\
Max repair rounds / patience & 10 / 3 & budget bound; three consecutive non-improvements ends R1 \\
Author cost / time cap & \$150 / 10 h & per campaign; exhaustion is reported as not compiled \\
Author session limit & 1 h & pilot sessions before the limit took 31--111 min; background work does not survive a session \\
Practice & $\le 30$ episodes per round; own simulator account & the author's means of in-session testing; logged \\
Exam & $n{=}100$, seeds from $[10^5,10^6)$ drawn after freezing & point-estimate threshold; interval half-width about \rWilsonHalfWidthNinetyFive{} points at the threshold; the interval is reported beside every verdict \\
Exam decision / completion & fail at $>5$ failures; always run to completion; one attempt & keeps failed campaigns comparable; no retake \\
Safety / speed / regression & 0 aborts; $\le 2\times$ expert median; not worse than any accepted head & a slow or regressed skill is not reusable \\
Settle before judging & 2 simulated s & an object released at the last moment is judged at rest \\
Caller check & 20 episodes, author product, 4 decisions $\times$ 8 steps & report only; warning if $>15$ points below exam \\
Re-evaluation (5.5.3) & 300 new layouts, exam version & report only; resolves abort rates the exam cannot \\
Held-out re-run (5.8.0) & every $R1.k$ and $C.k$ on the validation layouts & report only; the per-round comparison on unseen layouts \\
Second experiment (5.9.0, pre-registered 2026-10-10) & 150 development seeds (\rFEightDevNMin{} usable); inherit-only lineage $I.k$ (practice runs kept, no development-layout feedback), rounds matched to repair, own host account, not examined; gate, targets and limits unchanged & tests whether a block that shows 1--3\% aborts lets repair clear the gate, and separates inheritance from the development feedback \\
\protoB seeds & 50 per skill, exam range, excluding all manifest seeds & disjoint from every compile-time seed \\
\protoB caller & \texttt{deepseek-flash}, temperature 0, 32{,}768 output tokens, 12 decisions, 900 s per decision & temperature 0 (hosted decoding is not bit-reproducible); the decision cap is reported as a mean, not reached \\
\bottomrule
\end{tabular}
\end{table}

\noindent Before release each package must satisfy: (1)~the front matter validates and \texttt{atoms} matches \texttt{ATOMS}; (2)~both Python files pass the restricted-Python validator; (3)~every atom returns a dictionary with \texttt{ok}; (4)~\texttt{fallback(api)} completes without an uncaught error on every development layout; (5)~development and validation targets are met; (6)~the ledger chain, runtime manifest and official tree verify; (7)~exam, safety, speed and regression gates pass; (8)~files are hashed, the manifest is written, and the bundle is copied to the frozen store.

\section{Prompts}
\label{app:prompts}
The author prompt (\texttt{TASK.md}, about 12\,KB, identical for both products and every campaign) states the task, the four deliverables, the two run modes, the judging rule (fallback exam at $\ge 95\%$ on unseen layouts with zero safety aborts, speed and regression), the primitive API and its semantics, the per-episode limits and the settle period, the rules of the test (API-only access at run time; no modification of evaluator, runtime or ledgers; no external interpretation of results; only the author's own working directory and host account), the practice tool and, in the default condition, the host account. Each repair round adds a short note naming the round and the practice budget and pointing to the history directory with every earlier version, its feedback and its verdict. The caller prompt states the caller's role, the plan format (one to eight steps of atom call, observe or state), the decision limit, the shared call budget, the instruction to follow the standard flow and, on failure, the recovery section of \texttt{SKILL.md}, and the JSON answer format; it is identical for the caller check and for \protoB. In the write-once condition the practice paragraphs are replaced by one sentence stating that no simulator is available and the package cannot be run before delivery. Full prompt texts are released with the code.

\section{Price Snapshot}
\label{app:prices}
\begin{table}[H]
\centering
\caption{Prices per million tokens used for the caller ledger, captured 2026-10-08 from the provider's pricing page; every decision stores the price applicable at its UTC time.}
\small
\begin{tabular}{@{}p{3.5cm}p{2.3cm}cccp{4.4cm}@{}}
\toprule
Model & Provider & Input & Cached input & Output & Note \\
\midrule
\texttt{deepseek-flash} (V4.1-Flash), peak & DeepSeek & \$0.30 & \$0.006 & \$1.20 & UTC 01--04 and 06--10, weekdays \\
\texttt{deepseek-flash} (V4.1-Flash), off-peak & DeepSeek & \$0.15 & \$0.003 & \$0.60 & all other hours; Chinese public holidays are priced at peak (overstates cost) \\
Author (\rClaudeProduct) & Anthropic & \multicolumn{3}{c}{product-reported API-equivalent cost} & subscription channel; amount as reported by the product per session \\
Author (\rGPTProduct) & OpenAI & \multicolumn{3}{c}{not observable} & subscription channel; recorded as null, never estimated \\
\bottomrule
\end{tabular}
\end{table}

\section{Pre-registration of the Second Experiment and Deviation Record}
\label{app:prereg}
The following is a condensed translation of the registration written into the protocol on 2026-10-10, before any campaign of the second experiment ran, and of the deviation record appended during execution; the original text is released with the code.

\paragraph{Registered design.} Task \texttt{stack\_bowls\_three}, clutter tier, Claude Code as author, three campaigns. Development seeds 70000--70149 (150; the validation block 71000--71049 unchanged for comparability). Three lineages from the same R0 starting point with the same product, practice budget and 60-minute sessions: repair (unchanged), no-feedback rewrite (unchanged), and inherit-only, which copies the starting workspace with the author's notes and practice records, deletes the starting version's feedback and verdict, revises the previous version each round with the same practice budget, and receives none of the runtime's development-layout results for any version (no development evidence in the workspace or the host view; the task card and round prompt say so); the runtime keeps the best inherit-only version by the same strict-improvement rule and reports only \texttt{HEAD}. Rounds matched to repair, run in the background; own host account; not charged to the repair cost and time caps; not examined. Gate, ladder (ten rounds, patience three, 96\% and 94\% targets) and exam (100 episodes, $\geq 95\%$ and zero aborts) unchanged. Primary outcome: campaigns frozen out of three, descriptive. Secondary outcomes: RSI delta and exam aborts per campaign beside the six earlier campaigns; after each campaign, the starting version and every repair, control and inherit-only version run once on the hidden validation layouts, paired by round as repair$-$inherit, inherit$-$control and repair$-$control, with a one-sided sign test over campaigns and descriptive per-round counts; development-layout curves descriptive only. Registered limitations: the larger block and the new lineage are introduced together; the inherit-only author still learns whether \texttt{HEAD} changed; its host account is new; three campaigns on one task are existence evidence.

\paragraph{Deviations recorded during execution.}
\begin{itemize}
  \item 2026-10-11: the first three campaigns (i1--i3) were voided before any author session because the queue did not pass the new flags; they stopped in phase 0 and were re-queued unchanged as j1--j3.
  \item 2026-10-12 and 10-13/14: the SSH relay to the simulation host dropped several times; the coordinator was restarted each time with no author session in progress, existing episode results were kept and unfinished evaluations resumed under per-episode locks.
  \item 2026-10-13, about 03:50 UTC: every author session began failing within seconds with a provider authorization error; the runtime recorded the empty deliveries as non-improving rounds. At 06:10 UTC, before j3's exam score had been seen, the operator voided those rounds as infrastructure failures and returned each campaign to its state before the failure: j3's R1.7, R1.8, C.7 and C.8 and j2's I.6 were moved aside, not deleted, and the state change is recorded. j3 therefore resumed R1 after six repair rounds with one non-improving round; the exam that had been drawn when the failure stopped R1 was voided and j3's exam was redrawn when R1 ended (\texttt{R2\_exam\_2}); j2's three non-improving rounds were real sessions and its exam stood.
  \item 2026-10-14: j3's ledger had grown to 3{,}876 rows and its final report exceeded the coordinator's 600-second remote-command limit; an operator launcher that raises only that limit, with the runtime files and manifest untouched, produced the report and the held-out re-run.
  \item j2's isolation audit was graded review (another campaign's directory name and identifiers appeared in process listings; no content read); the operator reviewed the events on 2026-10-14 and retained the campaign.
\end{itemize}

\section{Result Files}
\label{app:files}
Every campaign directory (\texttt{data/protocol\_A/<campaign>/} in the released repository) holds fixed-schema files, one row per event, so that all tables can be rebuilt by one script: \texttt{campaign.json} (scene, seed ranges, limits, environment fingerprint, runtime manifest hash, official tree hash); \texttt{results/r2\_gate\_spec.json} (the locked gate); \texttt{results/episodes.jsonl} (hash-chained, one row per episode with stage, mode, version, seed, outcome, failure class, safety events, primitive calls, atom segments and coverage, decisions, times and cost); \texttt{results/\allowbreak measurement\_draws.jsonl} (exam, caller-check and re-evaluation draws); \texttt{state.json} (R0, repair log, control versions, validation checks); \texttt{summary.json} (outcome, compile cost, curves, gate checks, caller check, control, isolation audit); \texttt{held\_out\_*.json} and \texttt{extra\_eval\_*.json} (report-only re-runs); \texttt{phase0\_summary.json} (the health check and the expert runs). The frozen packages are released as \texttt{frozen\_skills/<task>\_<tag>/} (the three frozen files and the manifest). A reuse evaluation (\texttt{data/protocol\_B/<evaluation>/}) writes \texttt{reuse\_summary.json} and \texttt{decisions.jsonl} (one row per caller decision with tokens, price, latency and raw response) beside its ledger. The numbers in this paper are produced by one script (\texttt{analysis/paper/}) from these files and from the per-campaign analysis tables (\texttt{analysis/data/}); the published values in \cref{tab:published} come from a transcription file (\texttt{analysis/data/f09\_reference.json}, checked against the PDF of arXiv:2506.18088v2) and the phase-0 expert column from each campaign's \texttt{phase0\_summary.json}.

\end{document}

%% file: data/results_numbers.tex
\newcommand{\rnClaude}{20}

\newcommand{\rnCampaigns}{25}
\newcommand{\rnTasks}{7}
\newcommand{\rnFrozen}{7}
\newcommand{\rnFailedExam}{4}
\newcommand{\rnNotCompiled}{9}
\newcommand{\rnTasksFrozen}{4}
\newcommand{\rnRateBar}{12}

\newcommand{\rnFrozenZeroRepair}{5}

\newcommand{\rnRepaired}{14}

\newcommand{\rnControl}{10}

\newcommand{\rnControlComparable}{8}
\newcommand{\rFrozenUSDMedian}{6.69}
\newcommand{\rFrozenUSDMin}{4.29}
\newcommand{\rFrozenUSDMax}{8.22}

\newcommand{\rFrozenHoursMin}{1.4}
\newcommand{\rFrozenHoursMax}{2.9}

\newcommand{\rAllUSDTotal}{384}

\newcommand{\rAllHoursTotal}{78}
\newcommand{\rNCUSDMedian}{36.74}
\newcommand{\rNCUSDMin}{10.97}
\newcommand{\rNCUSDMax}{48.28}

\newcommand{\rFrozenWallMin}{5.2}
\newcommand{\rFrozenWallMax}{13.0}

\newcommand{\rSimEpisodesMedian}{532}
\newcommand{\rSimEpisodesMin}{262}
\newcommand{\rSimEpisodesMax}{1177}

\newcommand{\rFrozenSessionsMin}{2}
\newcommand{\rFrozenSessionsMax}{4}

\newcommand{\rDeltaMeanRepaired}{+13.3}
\newcommand{\rDeltaMedianRepaired}{+9.9}
\newcommand{\rDeltaMinRepaired}{-5.3}
\newcommand{\rDeltaMaxRepaired}{+39.4}
\newcommand{\rnDeltaPositive}{12}

\newcommand{\rBigDeltaList}{+39.4, +34.4, +28.5, +22.3}
\newcommand{\rCorrBaselineDelta}{-0.88}

\newcommand{\rExamMeanBaseLow}{88.9}
\newcommand{\rExamMeanBaseHigh}{92.1}

\newcommand{\rDevNMin}{42}
\newcommand{\rDevNMax}{47}
\newcommand{\rValNMin}{36}
\newcommand{\rValNMax}{46}

\newcommand{\rExamsStackBowlsThree}{78, 90, 86, 90, 89, 95}

\newcommand{\rExamsPlaceDualShoes}{78, 90, 88}

\newcommand{\rSBThreeUSDTotal}{202}
\newcommand{\rSafetyOnlyAbortsList}{1, 1, 3, 2, 4}
\newcommand{\rSafetyOnlyExamMin}{95}
\newcommand{\rSafetyOnlyExamMax}{97}
\newcommand{\rExamAbortsTotal}{75}
\newcommand{\rExamAbortsWorkspace}{61}
\newcommand{\rExamAbortsDropped}{14}

\newcommand{\rAbortUpperZero}{3.6}
\newcommand{\rAbortUpperOne}{5.4}

\newcommand{\rCCValid}{138}
\newcommand{\rCCSuccess}{137}
\newcommand{\rCCRate}{99.3}

\newcommand{\rCCDecisionsMin}{1.0}
\newcommand{\rCCDecisionsMax}{3.1}

\newcommand{\rPZeroOne}{37}
\newcommand{\rPZeroTwo}{13}
\newcommand{\rPZeroThree}{5}
\newcommand{\rIsoClean}{17}
\newcommand{\rIsoMetadata}{2}
\newcommand{\rIsoReview}{1}

\newcommand{\rSpeedRatioMedian}{1.18}
\newcommand{\rSpeedRatioMax}{1.49}
\newcommand{\rGPTExamMin}{83}
\newcommand{\rGPTExamMax}{96}
\newcommand{\rGPTAbortsList}{1, 1, 0, 9, 5}
\newcommand{\rGPTRateBar}{1}

\newcommand{\rGPTDeltaMin}{-1.7}
\newcommand{\rGPTDeltaMax}{+3.0}
\newcommand{\rGPTBaseMin}{80}
\newcommand{\rGPTBaseMax}{98}
\newcommand{\rGPTRepairsMin}{1}
\newcommand{\rGPTRepairsMax}{7}

\newcommand{\rExtraEvalN}{300}
\newcommand{\rHORounds}{34}
\newcommand{\rHOPos}{26}
\newcommand{\rHONeg}{7}
\newcommand{\rHOTie}{1}
\newcommand{\rHOMeanDiff}{+6.4}
\newcommand{\rHOCampaigns}{8}
\newcommand{\rHOCampPos}{8}

\newcommand{\rHOSignP}{0.004}
\newcommand{\rHOPermMean}{+5.7}
\newcommand{\rHOPermP}{0.004}
\newcommand{\rHOBestRepair}{5}
\newcommand{\rHOBestControl}{3}

\newcommand{\rHOLayoutsMin}{36}
\newcommand{\rHOLayoutsMax}{45}
\newcommand{\rHOEpisodesApprox}{3,200}

\newcommand{\rDevAllRounds}{34}
\newcommand{\rDevAllPos}{29}

\newcommand{\rDevAllMeanDiff}{+6.2}
\newcommand{\rHOControlMin}{4/45}
\newcommand{\rHOControlMax}{44/45}
\newcommand{\rMechSessions}{60}
\newcommand{\rMechReadFeedback}{60}
\newcommand{\rMechReadVerdict}{59}
\newcommand{\rMechReadNotes}{60}
\newcommand{\rMechCites}{56}
\newcommand{\rMechCitesPct}{93}

\newcommand{\rWODraftsHostOnceDiff}{+0.5}
\newcommand{\rWODraftsHostOnceP}{0.48}

\newcommand{\rWOBaseHostOnceDiff}{-3.4}
\newcommand{\rWOBaseHostOnceP}{0.77}

\newcommand{\rWOBaseNoHostOnceDiff}{-3.9}
\newcommand{\rWOBaseNoHostOnceP}{0.97}

\newcommand{\rWOOnceBaseMin}{76}
\newcommand{\rWOOnceBaseMax}{98}

\newcommand{\rReuseMoveCanPotODec}{1.0}

\newcommand{\rReusePlacePhoneStandODec}{2.0}

\newcommand{\rReuseStackBowlsTwoFAborts}{1}
\newcommand{\rReuseStackBowlsTwoOAborts}{2}

\newcommand{\rPoolF}{148/150}
\newcommand{\rPoolO}{147/150}
\newcommand{\rPoolFPct}{98.7}
\newcommand{\rPoolOPct}{98.0}
\newcommand{\rPoolFCI}{95.3--99.6}
\newcommand{\rPoolOCI}{94.3--99.3}

\newcommand{\rOdsTotalCost}{0.055}
\newcommand{\rOdsDecisions}{200}

\newcommand{\rOdsCostMin}{0.00028}
\newcommand{\rOdsCostMax}{0.00054}
\newcommand{\rOdsLatencyMedian}{2.0}
\newcommand{\rOdsLatencyPNinety}{2.8}

\newcommand{\rOdsInTokensMean}{3234}
\newcommand{\rOdsOutTokensMean}{336}
\newcommand{\rReuseCompileTotal}{20.20}
\newcommand{\rReuseCompileMin}{5.46}
\newcommand{\rReuseCompileMax}{7.99}
\newcommand{\rAmortThousandMax}{0.0083}
\newcommand{\rAmortHundredMax}{0.080}
\newcommand{\rSideDstN}{2}
\newcommand{\rSideDstSuccess}{0}
\newcommand{\rSideDstCost}{4.77}
\newcommand{\rSideDstDec}{9.5}
\newcommand{\rSideDdsN}{2}
\newcommand{\rSideDdsSuccess}{0}
\newcommand{\rSideDdsCost}{0.039}

\newcommand{\rSideDstOutTokens}{201}

\newcommand{\rYieldCampaigns}{25}
\newcommand{\rYieldEpisodesAll}{18,941}
\newcommand{\rYieldEpisodesValid}{18,764}
\newcommand{\rYieldEpisodesInfra}{177}
\newcommand{\rYieldInfraPct}{0.9}
\newcommand{\rYieldSuccess}{15,909}
\newcommand{\rYieldFail}{2,855}

\newcommand{\rYieldFailFinished}{1,684}
\newcommand{\rYieldFailWorkspace}{893}
\newcommand{\rYieldFailDropped}{254}

\newcommand{\rYieldSubgoalsSucc}{5.6}
\newcommand{\rYieldSubgoalsFail}{5.3}
\newcommand{\rYieldSegmentsMean}{8.4}

\newcommand{\rYieldCoverageFull}{99.2}
\newcommand{\rYieldEpisodesProtocol}{14,466}
\newcommand{\rYieldEpisodesPractice}{2,938}
\newcommand{\rYieldEpisodesPostHoc}{4,475}
\newcommand{\rYieldFailInvalidQuat}{2}
\newcommand{\rYieldFailAtomError}{21}
\newcommand{\rYieldFailBudget}{1}

\newcommand{\rYieldSlotHours}{366}
\newcommand{\rYieldPerSlotHour}{51.3}

\newcommand{\rYieldPerWallHour}{60.5}
\newcommand{\rYieldWallMedianSec}{70}
\newcommand{\rBDiscFonly}{1}
\newcommand{\rBDiscOonly}{0}
\newcommand{\rBBothFail}{2}

\newcommand{\rFJOneDevBest}{120/128}
\newcommand{\rFJOneUSD}{61.2}

\newcommand{\rFJTwoDevBest}{122/128}

\newcommand{\rFJThreeDevBest}{122/128}
\newcommand{\rFJThreeUSD}{119.7}

\newcommand{\rFEightRCMean}{+10.2}
\newcommand{\rFEightRCPos}{16}
\newcommand{\rFEightRCNeg}{1}
\newcommand{\rFEightRCTie}{1}

\newcommand{\rFEightRCSignP}{0.125}
\newcommand{\rFEightRCPerCamp}{+15.8, +6.3, +8.5}
\newcommand{\rFEightICMean}{+8.7}
\newcommand{\rFEightICPos}{14}
\newcommand{\rFEightICNeg}{3}
\newcommand{\rFEightICTie}{1}

\newcommand{\rFEightICPerCamp}{+12.5, +5.3, +8.4}
\newcommand{\rFEightRIMean}{+1.5}
\newcommand{\rFEightRIPos}{11}
\newcommand{\rFEightRINeg}{6}
\newcommand{\rFEightRITie}{1}
\newcommand{\rFEightRICampPos}{3}

\newcommand{\rFEightRISignP}{0.125}
\newcommand{\rFEightRIPerCamp}{+3.2, +1.0, +0.1}
\newcommand{\rFEightN}{3}
\newcommand{\rFEightFrozen}{0}
\newcommand{\rFEightExams}{88, 93, 94}
\newcommand{\rFEightAborts}{8, 5, 3}
\newcommand{\rFEightRoundsList}{4, 6, 8}
\newcommand{\rFEightDevNMin}{128}

\newcommand{\rFEightAbortWorkspace}{13}
\newcommand{\rFEightAbortDropped}{3}

\newcommand{\rFEightCostRepair}{104}
\newcommand{\rFEightCostControl}{78}
\newcommand{\rFEightCostInherit}{96}
\newcommand{\rFEightCostTotal}{278}
\newcommand{\rFEightWallMin}{47}
\newcommand{\rFEightWallMax}{74}

\newcommand{\rFEightDeltaList}{+1.3, +26.3, +20.7}
\newcommand{\rFEightHOLayouts}{45}
\newcommand{\rFEightDevAbortsExam}{7, 4, 4}
\newcommand{\rFEightDevAbortsStart}{6, 36, 16}
\newcommand{\rFEightDevAbortsMinRound}{4}
\newcommand{\rFEightDevAbortsStartMax}{36}
\newcommand{\rFEightControlMin}{4}
\newcommand{\rFEightControlMax}{42}
\newcommand{\rSBThreeAllCampaigns}{9}
\newcommand{\rSBThreeAllFrozen}{0}
\newcommand{\rSBThreeAllExamMin}{78}
\newcommand{\rSBThreeAllExamMax}{95}
\newcommand{\rSBThreeAllAbortsMin}{3}
\newcommand{\rSBThreeAllAbortsMax}{16}
\newcommand{\rHORepairAboveStart}{8}
\newcommand{\rHORepairStartMean}{+4.3}

\newcommand{\rHOControlBelowStart}{6}
\newcommand{\rHOControlStartMean}{-1.3}
\newcommand{\rHORepairStartSignP}{0.004}
\newcommand{\rHOMeanDiffPP}{+15.0}
\newcommand{\rLearnKList}{-2, +2, +1, +2, +0, +0, +7, +2}
\newcommand{\rLearnKPos}{5}
\newcommand{\rLearnKZero}{2}
\newcommand{\rLearnKNeg}{1}
\newcommand{\rCorrBaselineExam}{0.10}

\newcommand{\rCCScheduled}{140}
\newcommand{\rProgramUSDPerFrozen}{55}
\newcommand{\rProgramUSDPerTask}{96}

\newcommand{\rWilsonLBNinetyFive}{88.8}
\newcommand{\rWilsonHalfWidthNinetyFive}{4.5}
\newcommand{\rWOWithinSD}{11.8}
\newcommand{\rWOWithinDF}{13}
\newcommand{\rWOSE}{5.7}
\newcommand{\rWOMDE}{14}
\newcommand{\rAbortUpperOneTwenty}{3.0}

\newcommand{\rnHardAll}{18}
\newcommand{\rnFailedCampaignsClaude}{13}

\newcommand{\rFNinePZeroStackBowlsThree}{4--6}
\newcommand{\rFNinePZeroNStackBowlsThree}{12}
\newcommand{\rFNineEntries}{35}

\newcommand{\rFNineEasyBelowSixty}{24}
\newcommand{\rFNineHardBelowTwentyFive}{32}

\newcommand{\rFNineExamMin}{78}
\newcommand{\rFNineExamMax}{100}

\newcommand{\rFNineCodegenZero}{\texttt{place\_phone\_stand} and \texttt{hanging\_mug}}

\newcommand{\rFNinePZeroCampaigns}{43}
\newcommand{\rGPU}{NVIDIA GeForce RTX 5090}
\newcommand{\rDriver}{595.71.05}
\newcommand{\rTorch}{2.8.0+cu128}

\newcommand{\rRuntimeMin}{5.5.0}
\newcommand{\rRuntimeMax}{5.6.0}
\newcommand{\rGPTRuntimeMin}{5.6.1}
\newcommand{\rGPTRuntimeMax}{5.8.0}
\newcommand{\rClaudeProduct}{Claude Code 2.1.284}
\newcommand{\rClaudeModel}{claude-opus-5-5}
\newcommand{\rGPTProduct}{Codex CLI 0.160.0}
\newcommand{\rGPTModel}{gpt-6-astra}

%% file: data/tab_compile_rows.tex
\newcommand{\tabCompileRows}{%
\texttt{phone\_stand} & 1 & yes & 42/42 & 40/40 & 0 & 42 & 5.46 & 1.6 & 6.3 & 99 & 94.6--99.8 & 0 & -1.0 & frozen & 20/20 & clean \\
\texttt{phone\_stand} & 2 & yes & 30/42 & 27/40 & 2 & 30$\to$42 & 8.22 & 2.9 & 13.0 & 96 & 90.2--98.4 & 0 & +28.5 & frozen & 20/20 & clean \\
\texttt{phone\_stand} & 3 & -- & 42/42 & 39/40 & 0 & 42 & 7.97 & 1.7 & 5.8 & 95 & 88.8--97.8 & 1 & -2.5 & failed exam & -- & clean \\
\texttt{bowls\_two} & 1 & yes & 47/47 & 41/41 & 0 & 47 & 6.75 & 1.8 & 10.5 & 98 & 93.0--99.4 & 0 & -2.0 & frozen & 19/19 & clean \\
\texttt{bowls\_two} & 2 & yes & 41/47 & 39/41 & 1 & 41$\to$46 & 9.29 & 2.6 & 8.3 & 95 & 88.8--97.8 & 2 & -0.1 & failed exam & -- & clean \\
\texttt{bowls\_two} & 3 & -- & 44/47 & 38/41 & 4 & 44$\to$47 & 25.53 & 5.3 & 14.2 & 96 & 90.2--98.4 & 4 & +3.3 & failed exam & -- & clean \\
\texttt{can\_pot} & 1 & -- & 44/45 & 43/44 & 0 & 44 & 4.29 & 1.4 & 10.3 & 98 & 93.0--99.4 & 0 & +0.3 & frozen & 19/19 & clean \\
\texttt{can\_pot} & 2 & -- & 39/45 & 39/44 & 1 & 39$\to$44 & 6.12 & 1.5 & 8.0 & 99 & 94.6--99.8 & 0 & +10.4 & frozen & 19/20 & clean \\
\texttt{can\_pot} & 3 & -- & 45/45 & 44/44 & 0 & 45 & 7.99 & 1.8 & 6.7 & 100 & 96.3--100.0 & 0 & +0.0 & frozen & 20/20 & clean \\
\texttt{burger\_fries} & 1 & -- & 43/43 & 46/46 & 0 & 43 & 6.69 & 1.7 & 5.2 & 100 & 96.3--100.0 & 0 & +0.0 & frozen & 20/20 & clean \\
\texttt{hanging\_mug} & 1 & -- & 38/42 & 36/43 & 1 & 38$\to$42 & 12.77 & 2.6 & 6.8 & 97 & 91.5--99.0 & 1 & +13.3 & failed exam & -- & clean \\
\texttt{dual\_shoes} & 1 & yes & 42/44 & 30/36 & 3 & 42$\to$42 & 22.41 & 4.2 & 15.1 & 78 & 68.9--85.0 & 0 & -5.3 & not compiled & -- & clean \\
\texttt{dual\_shoes} & 2 & yes & 39/44 & 29/36 & 5 & 39$\to$42 & 36.74 & 5.7 & 16.6 & 90 & 82.6--94.5 & 1 & +9.4 & not compiled & -- & clean \\
\texttt{dual\_shoes} & 3 & -- & 38/44 & 24/36 & 4 & 38$\to$42 & 21.43 & 3.8 & 10.4 & 88 & 80.2--93.0 & 1 & +21.3 & not compiled & -- & clean \\
\texttt{bowls\_three} & 1 & yes & 38/43 & 35/45 & 3 & 38$\to$38 & 10.97 & 4.0 & 14.6 & 78 & 68.9--85.0 & 16 & +0.2 & not compiled & -- & metadata \\
\texttt{bowls\_three} & 2 & yes & 34/43 & 25/45 & 5 & 34$\to$42 & 17.67 & 5.4 & 19.6 & 90 & 82.6--94.5 & 9 & +34.4 & not compiled & -- & clean \\
\texttt{bowls\_three} & 3 & -- & 34/43 & 35/45 & 6 & 34$\to$42 & 41.35 & 6.8 & 17.0 & 86 & 77.9--91.5 & 9 & +8.2 & not compiled & -- & clean \\
\texttt{bowls\_three} & 4 & yes & 33/43 & 40/45 & 7 & 33$\to$40 & 38.16 & 6.1 & 22.9 & 90 & 82.6--94.5 & 8 & +1.1 & not compiled & -- & clean \\
\texttt{bowls\_three} & 5 & yes & 31/43 & 30/45 & 8 & 31$\to$42 & 45.92 & 8.2 & 28.8 & 89 & 81.4--93.7 & 4 & +22.3 & not compiled & -- & metadata \\
\texttt{bowls\_three} & 6 & -- & 27/43 & 25/45 & 10 & 27$\to$43 & 48.28 & 9.1 & 23.2 & 95 & 88.8--97.8 & 3 & +39.4 & not compiled & -- & review \\
\midrule
\texttt{hanging\_mug} & 1 & -- & 38/42 & 42/43 & 1 & 38$\to$41 & -- & 2.2 & 6.0 & 96 & 90.2--98.4 & 1 & -1.7 & failed exam & -- & clean \\
\texttt{dual\_shoes} & 1 & -- & 39/44 & 33/36 & 7 & 39$\to$43 & -- & 8.4 & 29.6 & 94 & 87.5--97.2 & 1 & +2.3 & not compiled & -- & clean \\
\texttt{dual\_shoes} & 2 & -- & 42/44 & 33/36 & 3 & 42$\to$42 & -- & 4.9 & 19.1 & 93 & 86.3--96.6 & 0 & +1.3 & not compiled & -- & clean \\
\texttt{bowls\_three} & 1 & -- & 39/43 & 38/45 & 4 & 39$\to$41 & -- & 5.8 & 14.3 & 83 & 74.5--89.1 & 9 & -1.4 & not compiled & -- & clean \\
\texttt{bowls\_three} & 2 & -- & 37/43 & 36/45 & 6 & 37$\to$38 & -- & 7.7 & 42.2 & 83 & 74.5--89.1 & 5 & +3.0 & not compiled & -- & clean \\}

%% file: data/tab_extra_eval_rows.tex
\newcommand{\tabExtraEvalRows}{%
\texttt{phone\_stand} 3 & 95 & 1 & 290/300 & 94.0--98.2 & 0 & 0.0--1.2 \\
\texttt{hanging\_mug} 1 & 97 & 1 & 290/300 & 94.0--98.2 & 3 & 0.2--2.9 \\
\texttt{bowls\_two} 3 & 96 & 4 & 288/300 & 93.1--97.7 & 8 & 1.2--5.2 \\
\texttt{bowls\_two} 2 & 95 & 2 & 288/300 & 93.1--97.7 & 11 & 1.8--6.5 \\}

%% file: data/tab_heldout_rows.tex
\newcommand{\tabHeldoutRows}{%
\texttt{bowls\_three} & c7a1f0 & 33/45 & 3 & 36/27, 33/33, 35/43 & +0.3 & 36 / 43 \\
\texttt{bowls\_three} & ef1eb1 & 31/45 & 5 & 40/24, 41/38, 34/37, 42/21, 36/23 & +10.0 & 42 / 38 \\
\texttt{phone\_stand} & 4161b3 & 28/40 & 2 & 34/25, 39/38 & +5.0 & 39 / 38 \\
\texttt{bowls\_two} & 102fc5 & 38/41 & 1 & 40/34 & +6.0 & 40 / 34 \\
\texttt{bowls\_three} & 451bca & 38/45 & 7 & 36/41, 40/26, 38/36, 39/42, 40/39, 39/33, 37/39 & +1.9 & 40 / 42 \\
\texttt{dual\_shoes} & 320e4b & 32/36 & 3 & 35/31, 31/25, 32/28 & +4.7 & 35 / 31 \\
\texttt{bowls\_three} & 1cad2d & 28/45 & 8 & 38/27, 35/37, 40/4, 38/32, 43/44, 38/23, 41/37, 43/18 & +11.8 & 43 / 44 \\
\texttt{dual\_shoes} & 4e70ca & 31/36 & 5 & 33/30, 35/26, 32/31, 33/29, 34/22 & +5.8 & 35 / 31 \\}

%% file: data/tab_writeonce_rows.tex
\newcommand{\tabWriteonceRows}{%
\texttt{phone\_stand} & 88.3 (3) & 85.0 (2) & 92.5 (1) \\
\texttt{bowls\_two} & 95.9 (3) & 96.3 (2) & 97.6 (1) \\
\texttt{can\_pot} & 95.5 (3) & 96.6 (2) & 95.5 (1) \\
\texttt{hanging\_mug} & 83.7 (1) & 75.6 (2) & 83.7 (1) \\
\texttt{dual\_shoes} & 76.9 (3) & -- & 86.1 (1) \\
\texttt{bowls\_three} & 70.4 (6) & -- & 75.6 (1) \\}

%% file: data/tab_reuse_rows.tex
\newcommand{\tabReuseRows}{%
\texttt{can\_pot} & 7.99 & 100/100 & F & 50/50 & 92.9--100.0 & 0 & 0.0 & 0 & 0.080 & 0.0080 \\
 &  &  & O-ds & 50/50 & 92.9--100.0 & 0 & 1.0 & 0.00028 & 0.080 & 0.0083 \\
\texttt{phone\_stand} & 5.46 & 99/100 & F & 49/50 & 89.5--99.6 & 0 & 0.0 & 0 & 0.055 & 0.0055 \\
 &  &  & O-ds & 49/50 & 89.5--99.6 & 0 & 2.0 & 0.00054 & 0.055 & 0.0060 \\
\texttt{bowls\_two} & 6.75 & 98/100 & F & 49/50 & 89.5--99.6 & 1 & 0.0 & 0 & 0.068 & 0.0068 \\
 &  &  & O-ds & 48/50 & 86.5--98.9 & 2 & 1.0 & 0.00029 & 0.068 & 0.0070 \\}

%% file: data/tab_f08_rows.tex
\newcommand{\tabFEightRows}{%
j1 & 4 & 120/128 & 39/45 & 26.3 / 16.5 / 18.5 & 47 & 88 & 80.2--93.0 & 8 & +1.3 & not compiled & clean \\
j2 & 6 & 122/128 & 30/45 & 36.0 / 27.6 / 33.6 & 62 & 93 & 86.3--96.6 & 5 & +26.3 & not compiled & review \\
j3 & 8 & 122/128 & 33/45 & 41.5 / 34.2 / 44.0 & 74 & 94$^{\dagger}$ & 87.5--97.2 & 3 & +20.7 & not compiled & clean \\}

%% file: data/tab_f08_heldout_rows.tex
\newcommand{\tabFEightHeldoutRows}{%
j1 & 39 & 41, 43, 43, 38 & 41, 39, 32, 40 & 29, 4, 34, 35 & +15.8 & +12.5 & +3.2 \\
j2 & 31 & 38, 41, 40, 39, 40, 43 & 40, 40, 39, 37, 39, 40 & 30, 30, 39, 42, 27, 35 & +6.3 & +5.3 & +1.0 \\
j3 & 36 & 38, 39, 43, 42, 43, 39, 40, 40 & 31, 42, 42, 40, 42, 43, 41, 42 & 37, 23, 41, 35, 27, 39, 21, 33 & +8.5 & +8.4 & +0.1 \\}

%% file: data/tab_f08_compare_rows.tex
\newcommand{\tabFEightCompareRows}{%
repair $-$ control & inheritance and feedback together & +15.8, +6.3, +8.5 & +10.2 & 16/1/1 & 3/0 & 0.125 \\
inherit $-$ control & inheritance alone & +12.5, +5.3, +8.4 & +8.7 & 14/3/1 & 3/0 & 0.125 \\
repair $-$ inherit & feedback alone & +3.2, +1.0, +0.1 & +1.5 & 11/6/1 & 3/0 & 0.125 \\}

%% file: data/tab_f09_rows.tex
\newcommand{\tabFNineRows}{%
place\_phone\_stand & 15 & 35 & 2 & 13 & 44 & 6 & 7 & 0 & 0 & 2 & 0 & 66 & 8 (6) & 99, 96, 95 (2/3) \\
stack\_bowls\_two & 76 & 91 & 82 & 61 & 83 & 30 & 41 & 0 & 0 & 6 & 30 & 78 & 6--8 (6) & 98, 95, 96 (1/3) \\
move\_can\_pot & 25 & 58 & 22 & 39 & 70 & 12 & 21 & 4 & 0 & 6 & 30 & 93 & 9 (6) & 98, 99, 100 (3/3) \\
place\_burger\_fries & 50 & 80 & 49 & 72 & 72 & 27 & 4 & 0 & 0 & 18 & 100 & 97 & 10 (1) & 100 (1/1) \\
hanging\_mug & 23 & 11 & 7 & 8 & 17 & 16 & 3 & 0 & 0 & 1 & 0 & 63 & 9--10 (5) & 97 (0/1); GPT-6 96 \\
place\_dual\_shoes & 4 & 15 & 9 & 8 & 13 & 4 & 0 & 0 & 0 & 0 & 22 & 77 & 6--9 (7) & 78, 90, 88 (0/3); GPT-6 94, 93 \\
stack\_bowls\_three & 51 & 66 & 48 & 63 & 57 & 17 & 24 & 0 & 0 & 5 & 20 & 43 & 4--6 (12) & 78, 90, 86, 90, 89, 95 (0/6); GPT-6 83, 83; 150 dev.\ 88, 93, 94 \\}

%% file: figures/overview.tex
\begin{tikzpicture}[
  font=\small,
  box/.style={draw=RSIInk!45, rounded corners=3pt, align=center, inner sep=5pt, line width=0.5pt, fill=white},
  stage/.style={box, minimum height=1.15cm, text width=3.05cm},
  art/.style={box, fill=RSIPurple!8, draw=RSIPurple!70, line width=0.8pt, text width=3.7cm, minimum height=2.0cm},
  kern/.style={box, fill=RSIGray, text width=13.6cm},
  lab/.style={font=\footnotesize\bfseries, text=RSIInk!85},
  arr/.style={-{Latex[length=2.2mm]}, line width=0.7pt, draw=RSIInk!70},
  every node/.style={text=RSIInk}]

  \node[stage, fill=RSIBlue!7, draw=RSIBlue!70] (r0) at (0,0) {\textbf{R0 generate}\\[-1pt]{\scriptsize two independent drafts from the task card; keep the better; hidden-validation baseline}};
  \node[stage, fill=RSIBlue!7, draw=RSIBlue!70] (r1) at (3.9,0) {\textbf{R1 repair}\\[-1pt]{\scriptsize new session per round: read earlier versions' dev-layout evidence, revise; strictly better replaces the head}};
  \node[stage, fill=RSIGreen!8, draw=RSIGreen!70] (r2) at (8.05,0) {\textbf{R2 qualify}\\[-1pt]{\scriptsize 100 exam layouts drawn after freezing: $\geq 95\%$, zero safety aborts, speed, no regression}};
  \node[art] (fs) at (12.45,0) {\textbf{FrozenSkill}\\[1pt]{\scriptsize\texttt{SKILL.md}, \texttt{skills.py}, \texttt{perception.py}}\\[-1pt]{\scriptsize\texttt{MANIFEST.json}: hashes, seeds, environment fingerprint, exam outcome, compile cost}\\[1pt]{\scriptsize immutable after release}};

  \draw[arr] (r0) -- (r1);
  \draw[arr] (r1) -- (r2);
  \node[diamond, draw=RSIGreen!80, fill=RSIGreen!25, inner sep=1.6pt] (gate) at ($(r1.east)!0.5!(r2.west)$) {};
  \node[font=\scriptsize, text=RSIInk!80, align=center, anchor=south] (val) at ($(gate.north)+(0,0.72)$) {hidden validation layouts\\(never fed back; dev $\geq 96\%$ and val $\geq 94\%$ end R1)};
  \draw[RSIInk!60, line width=0.5pt] (val.south) -- (gate.north);
  \draw[arr] (r2) -- (fs);
  \node[lab, anchor=south west] at ($(r0.north west)+(0,1.3)$) {Protocol A: compile offline with an isolated, audited strong author (frozen weights; no model at run time)};
  \node[font=\scriptsize, text=RSIOrange!90!black, align=center, anchor=north] at ($(r1.south)+(0,-0.1)$) {no-feedback control draft $C.k$ beside every repair round; every $R1.k$ and $C.k$ re-run afterwards on the hidden layouts};

  \node[stage, fill=RSIOrange!9, draw=RSIOrange!80, text width=3.7cm] (exec) at (12.45,-3.05) {\textbf{R3 reuse}\\[-1pt]{\scriptsize same frozen files; 50 new layouts per skill, disjoint from every compile-time seed}};
  \node[box, text width=3.05cm] (arms) at (8.05,-3.05) {\scriptsize\textbf{executors}\\ F: \texttt{fallback(api)}, no model, \$0\\ O-ds: DeepSeek-flash caller,\\ reads \texttt{SKILL.md} only};
  \node[box, text width=3.05cm] (checks) at (3.9,-3.05) {\scriptsize\textbf{post-freeze checks}\\ caller check (20 layouts, author product)\\ 300-layout re-evaluation of near-bar skills};
  \node[box, text width=3.05cm] (out) at (0,-3.05) {\scriptsize\textbf{reported}\\ success with Wilson CI, safety aborts,\\ decisions, cost per episode,\\ amortized cost per deployment};

  \draw[arr] (fs) -- (exec);
  \draw[arr] (exec) -- (arms);
  \draw[arr] (arms) -- (checks);
  \draw[arr] (checks) -- (out);
  \node[lab, anchor=north west] at ($(out.south west)+(0,-0.12)$) {Protocol B: run cheap, measure reuse (skill never modified; callers read only \texttt{SKILL.md})};

  \node[kern, anchor=north, align=center] at ($(checks.south)+(2.3,-0.95)$) {\scriptsize\textbf{Shared, non-bypassable runtime} (identical for compile and reuse): primitive API \{observe, ee\_pose, gripper\_state, move\_pose, gripper, wait\} $\cdot$ tested IK and motion planning $\cdot$ workspace, joint and drop monitors with safety abort $\cdot$ restricted-Python skills $\cdot$ hash-chained ledger $\cdot$ environment fingerprint $\cdot$ author isolation and session audit};
\end{tikzpicture}

%% file: figures/learning.tex
\begin{tikzpicture}
\begin{groupplot}[
  group style={group size=2 by 1, horizontal sep=1.5cm},
  width=0.45\textwidth, height=6.0cm,
  ylabel style={font=\footnotesize}, xlabel style={font=\footnotesize},
  tick label style={font=\footnotesize}, title style={font=\footnotesize, yshift=-3pt},
  grid=major, grid style={RSIInk!12},
  legend style={font=\scriptsize, draw=none, fill=white, fill opacity=0.85, text opacity=1, row sep=-3pt, inner sep=1pt},
  legend cell align=left,
]
\nextgroupplot[title={(a) repair vs.\ rewrite, hidden layouts}, xlabel={control $C.k$ (\% success)}, ylabel={repair $R1.k$ (\% success)},
  xmin=0, xmax=100, ymin=45, ymax=100, legend pos=south east, legend columns=1]
\addplot[RSIInk!40, thin, dashed, domain=0:100, forget plot] {x};
\addplot[only marks, mark=*, mark size=2.00pt, RSIBlue, discard if not={task}{stack_bowls_three}]
  table[x expr=100*\thisrow{control_rate}, y expr=100*\thisrow{repair_rate}, col sep=comma]{data/heldout_pairs.csv};
\addlegendentry{bowls\_three (4 runs)}
\addplot[only marks, mark=square*, mark size=1.88pt, RSIOrange!90!black, discard if not={task}{place_dual_shoes}]
  table[x expr=100*\thisrow{control_rate}, y expr=100*\thisrow{repair_rate}, col sep=comma]{data/heldout_pairs.csv};
\addlegendentry{dual\_shoes (2)}
\addplot[only marks, mark=triangle*, mark size=2.50pt, RSIGreen!80!black, discard if not={task}{place_phone_stand}]
  table[x expr=100*\thisrow{control_rate}, y expr=100*\thisrow{repair_rate}, col sep=comma]{data/heldout_pairs.csv};
\addlegendentry{phone\_stand (1)}
\addplot[only marks, mark=diamond*, mark size=2.50pt, RSIPurple, discard if not={task}{stack_bowls_two}]
  table[x expr=100*\thisrow{control_rate}, y expr=100*\thisrow{repair_rate}, col sep=comma]{data/heldout_pairs.csv};
\addlegendentry{bowls\_two (1)}

\nextgroupplot[title={(b) exam vs.\ validation baseline}, xlabel={validation baseline before repair (\%)}, ylabel={exam success (\% of 100)},
  xmin=38, xmax=102, ymin=74, ymax=102, legend pos=south west, legend columns=1,
  extra y ticks={95}, extra y tick labels={}, extra y tick style={grid style={RSIGreen!60!black, dotted, thin}}]
\addplot[RSIInk!40, thin, dashed, domain=38:102, forget plot] {x};
\addplot[only marks, mark=*, mark size=2.25pt, RSIBlue, discard if not={status}{frozen}]
  table[x=baseline, y=exam, col sep=comma]{data/exam_vs_baseline.csv};
\addlegendentry{Claude, frozen}
\addplot[only marks, mark=o, mark size=2.25pt, RSIBlue, thick, discard if not={status}{failed_exam}, discard if not={agent}{claude}]
  table[x=baseline, y=exam, col sep=comma]{data/exam_vs_baseline.csv};
\addlegendentry{Claude, failed exam}
\addplot[only marks, mark=o, mark size=2.25pt, RSIRed, thick, discard if not={status}{not_compiled}, discard if not={agent}{claude}]
  table[x=baseline, y=exam, col sep=comma]{data/exam_vs_baseline.csv};
\addlegendentry{Claude, not compiled}
\addplot[only marks, mark=triangle, mark size=2.75pt, RSIOrange!90!black, thick, discard if not={agent}{gpt6}]
  table[x=baseline, y=exam, col sep=comma]{data/exam_vs_baseline.csv};
\addlegendentry{GPT-6 (all)}
\addplot[only marks, mark=diamond, mark size=2.75pt, RSIPurple, thick, discard if not={agent}{claude150}]
  table[x=baseline, y=exam, col sep=comma]{data/exam_vs_baseline.csv};
\addlegendentry{Claude, 150 dev layouts}
\end{groupplot}
\end{tikzpicture}

%% file: figures/inherit.tex
\begin{tikzpicture}
\begin{groupplot}[
  group style={group size=3 by 1, horizontal sep=0.8cm, ylabels at=edge left, yticklabels at=edge left},
  width=0.31\textwidth, height=5.4cm,
  ymin=0, ymax=47, ytick={0,15,30,45},
  ylabel style={font=\footnotesize}, xlabel style={font=\footnotesize},
  tick label style={font=\footnotesize}, title style={font=\footnotesize, yshift=-3pt},
  xlabel={round $k$}, grid=major, grid style={RSIInk!12},
  legend style={font=\scriptsize, draw=none, fill=white, fill opacity=0.85, text opacity=1, row sep=-3pt, inner sep=1pt},
  legend cell align=left,
]
\nextgroupplot[title={j1 (4 rounds)}, ylabel={successes / 45}, xmin=0.6, xmax=4.4, xtick={1,2,3,4}]
\addplot[RSIInk!60, thin, dashed, discard if not={run}{j1}] table[x=k, y=start, col sep=comma]{data/f08_heldout.csv};
\addplot[RSIBlue, thick, mark=*, mark size=1.88pt, discard if not={run}{j1}] table[x=k, y=repair, col sep=comma]{data/f08_heldout.csv};
\addplot[RSIPurple, thick, densely dashed, mark=square*, mark size=1.75pt, discard if not={run}{j1}] table[x=k, y=inherit, col sep=comma]{data/f08_heldout.csv};
\addplot[RSIOrange!90!black, thick, dotted, mark=triangle*, mark size=2.38pt, discard if not={run}{j1}] table[x=k, y=control, col sep=comma]{data/f08_heldout.csv};
\nextgroupplot[title={j2 (6 rounds)}, xmin=0.6, xmax=6.4, xtick={1,...,6}, legend pos=south west]
\addplot[RSIInk!60, thin, dashed, forget plot, discard if not={run}{j2}] table[x=k, y=start, col sep=comma]{data/f08_heldout.csv};
\addplot[RSIBlue, thick, mark=*, mark size=1.88pt, discard if not={run}{j2}] table[x=k, y=repair, col sep=comma]{data/f08_heldout.csv};
\addlegendentry{repair}
\addplot[RSIPurple, thick, densely dashed, mark=square*, mark size=1.75pt, discard if not={run}{j2}] table[x=k, y=inherit, col sep=comma]{data/f08_heldout.csv};
\addlegendentry{inherit only}
\addplot[RSIOrange!90!black, thick, dotted, mark=triangle*, mark size=2.38pt, discard if not={run}{j2}] table[x=k, y=control, col sep=comma]{data/f08_heldout.csv};
\addlegendentry{rewrite}
\nextgroupplot[title={j3 (8 rounds)}, xmin=0.6, xmax=8.4, xtick={1,...,8}, legend pos=south west]
\addplot[RSIInk!60, thin, dashed, discard if not={run}{j3}] table[x=k, y=start, col sep=comma]{data/f08_heldout.csv};
\addlegendentry{start (R0)}
\addplot[RSIBlue, thick, mark=*, mark size=1.88pt, forget plot, discard if not={run}{j3}] table[x=k, y=repair, col sep=comma]{data/f08_heldout.csv};
\addplot[RSIPurple, thick, densely dashed, mark=square*, mark size=1.75pt, forget plot, discard if not={run}{j3}] table[x=k, y=inherit, col sep=comma]{data/f08_heldout.csv};
\addplot[RSIOrange!90!black, thick, dotted, mark=triangle*, mark size=2.38pt, forget plot, discard if not={run}{j3}] table[x=k, y=control, col sep=comma]{data/f08_heldout.csv};
\end{groupplot}
\end{tikzpicture}

%% file: figures/reuse.tex
\begin{tikzpicture}
\begin{groupplot}[
  group style={group size=2 by 1, horizontal sep=1.5cm},
  width=0.45\textwidth, height=6.0cm,
  ylabel style={font=\footnotesize}, xlabel style={font=\footnotesize},
  tick label style={font=\footnotesize}, title style={font=\footnotesize, yshift=-3pt},
  grid=major, grid style={RSIInk!12},
  legend style={font=\scriptsize, draw=none, fill=white, fill opacity=0.85, text opacity=1, row sep=-3pt, inner sep=1pt},
  legend cell align=left,
]
\nextgroupplot[title={(a) success, Wilson 95\% CI}, ylabel={success (\%)},
  xmin=0.45, xmax=3.55, ymin=78, ymax=101.5, ytick={80,85,90,95,100},
  x tick label style={font=\scriptsize, rotate=18, anchor=north east, xshift=2pt},
  xtick={1,2,3}, xticklabels={can\_pot, phone\_stand, bowls\_two},
  legend pos=south west, legend columns=1,
  extra y ticks={95}, extra y tick labels={}, extra y tick style={grid style={RSIGreen!60!black, dotted, thin}}]
\addplot[only marks, mark=square*, mark size=2.12pt, RSIInk!70, discard if not={series}{exam},
  error bars/.cd, y dir=both, y explicit, error bar style={RSIInk!50}]
  table[x=x, y=rate, y error plus expr=\thisrow{hi}-\thisrow{rate}, y error minus expr=\thisrow{rate}-\thisrow{lo}, col sep=comma]{data/reuse_points.csv};
\addlegendentry{exam, author's gate ($n{=}100$)}
\addplot[only marks, mark=*, mark size=2.38pt, RSIBlue, discard if not={series}{F},
  error bars/.cd, y dir=both, y explicit, error bar style={RSIBlue!70}]
  table[x=x, y=rate, y error plus expr=\thisrow{hi}-\thisrow{rate}, y error minus expr=\thisrow{rate}-\thisrow{lo}, col sep=comma]{data/reuse_points.csv};
\addlegendentry{F: code only ($n{=}50$)}
\addplot[only marks, mark=triangle*, mark size=2.88pt, RSIOrange!90!black, discard if not={series}{O-ds},
  error bars/.cd, y dir=both, y explicit, error bar style={RSIOrange!70}]
  table[x=x, y=rate, y error plus expr=\thisrow{hi}-\thisrow{rate}, y error minus expr=\thisrow{rate}-\thisrow{lo}, col sep=comma]{data/reuse_points.csv};
\addlegendentry{O-ds: DeepSeek caller ($n{=}50$)}

\nextgroupplot[title={(b) cost per episode with compilation amortized}, xlabel={deployments $N$}, ylabel={USD per episode (log)},
  xmode=log, ymode=log, xmin=1, xmax=10000, ymin=0.0001, ymax=100,
  legend pos=south west, legend columns=1, clip=false]
\addplot[RSIBlue, thick, discard if not={skill}{move_can_pot}, discard if not={arm}{F}] table[x=N, y=per_episode, col sep=comma]{data/amortized.csv};
\addlegendentry{F, three skills}
\addplot[RSIBlue, thick, forget plot, discard if not={skill}{place_phone_stand}, discard if not={arm}{F}] table[x=N, y=per_episode, col sep=comma]{data/amortized.csv};
\addplot[RSIBlue, thick, forget plot, discard if not={skill}{stack_bowls_two}, discard if not={arm}{F}] table[x=N, y=per_episode, col sep=comma]{data/amortized.csv};
\addplot[RSIOrange!90!black, thick, densely dashed, discard if not={skill}{move_can_pot}, discard if not={arm}{O-ds}] table[x=N, y=per_episode, col sep=comma]{data/amortized.csv};
\addlegendentry{O-ds, three skills}
\addplot[RSIOrange!90!black, thick, densely dashed, forget plot, discard if not={skill}{place_phone_stand}, discard if not={arm}{O-ds}] table[x=N, y=per_episode, col sep=comma]{data/amortized.csv};
\addplot[RSIOrange!90!black, thick, densely dashed, forget plot, discard if not={skill}{stack_bowls_two}, discard if not={arm}{O-ds}] table[x=N, y=per_episode, col sep=comma]{data/amortized.csv};
\draw[RSIRed!80, thin, dotted] (axis cs:1,4.77) -- (axis cs:10000,4.77);
\draw[RSIRed!80, thin, dotted] (axis cs:1,0.039) -- (axis cs:10000,0.039);
\addlegendimage{RSIRed!80, thin, dotted}
\addlegendentry{no skill, self-test ($n{=}2$ each)}
\end{groupplot}
\end{tikzpicture}